\documentclass[11pt, letterpaper]{article}

\usepackage{amsmath}
\usepackage{amssymb}
\usepackage{amsthm}
\usepackage{graphicx}
\usepackage{hyperref}
\usepackage{url}
\usepackage{algorithm}
\usepackage{algorithmic}
\usepackage{caption}
\usepackage{subcaption}
\usepackage{booktabs}
\usepackage{authblk}
\usepackage{appendix}
\usepackage[utf8]{inputenc}
\usepackage[T1]{fontenc}
\usepackage[english]{babel}
\usepackage{geometry}
\usepackage{xcolor}
\usepackage{fancyhdr}
\usepackage{microtype}
\usepackage{parskip}
\usepackage{tabularx}
\usepackage{enumitem}

\hypersetup{
    colorlinks=true,
    linkcolor=blue,
    filecolor=magenta,      
    urlcolor=cyan,
    citecolor=red,
}

\definecolor{headercolor}{RGB}{0, 50, 100}
\title{Transformation vs Tradition: Artificial General Intelligence (AGI) for Arts and Humanities}

\author[1]{Zhengliang Liu$^*$}
\author[1]{Yiwei Li$^*$}
\author[2]{Qian Cao$^*$}
\author[3]{Junwen Chen$^*$}
\author[1]{Tianze Yang}
\author[1]{Zihao Wu}
\author[4]{John Gibbs}
\author[1]{Khaled Rasheed}
\author[1]{Ninghao Liu\textsuperscript{\textdagger}}
\author[2]{Gengchen Mai\textsuperscript{\textdagger}}
\author[1]{Tianming Liu\textsuperscript{\textdagger}}

\affil[1]{School of Computing, University of Georgia}
\affil[2]{Department of Geography, University of Georgia}
\affil[3]{Department of Computer Science and Engineering, Michigan State University}
\affil[4]{Department of Theatre \& Film Studies, University of Georgia}

\date{}

\begin{document}

\maketitle
\def\thefootnote{*}\footnotetext{These authors contributed equally to this work}
\def\thefootnote{\textsuperscript{\textdagger}}\footnotetext{Corresponding author(s). E-mail(s): ninghao.liu@uga.edu, gengchen.mai25@uga.edu, tliu@uga.edu}

\begin{abstract}
Recent advances in artificial general intelligence (AGI), particularly large language models and creative image generation systems have demonstrated impressive capabilities on diverse tasks spanning the arts and humanities. However, the swift evolution of AGI has also raised critical questions about its responsible deployment in these culturally significant domains traditionally seen as profoundly human. This paper provides a comprehensive analysis of the applications and implications of AGI for text, graphics, audio, and video pertaining to arts and the humanities. We survey cutting-edge systems and their usage in areas ranging from poetry to history, marketing to film, and communication to classical art. We outline substantial concerns pertaining to factuality, toxicity, biases, and public safety in AGI systems, and propose mitigation strategies. The paper argues for multi-stakeholder collaboration to ensure AGI promotes creativity, knowledge, and cultural values without undermining truth or human dignity. Our timely contribution summarizes a rapidly developing field, highlighting promising directions while advocating for responsible progress centering on human flourishing. The analysis lays the groundwork for further research on aligning AGI's technological capacities with enduring social goods.

\end{abstract}

\section{Introduction}

% The swift evolution of AI and its impact on art/humanity domains.

% For example, NightCafe Studio users have generated more than 75 million images and DALL-E 2’s 1.5 million users are generating more than 2 million images daily~\footnote{\url{https://venturebeat.com/ai/what-you-need-to-know-about-ai-generated-art/}}.

% "Traditional categories within the arts~\footnote{\url{https://www.britannica.com/topic/the-arts}} include literature (including poetry, drama, story, and so on), the visual arts (painting, drawing, sculpture, etc.), the graphic arts (painting, drawing, design, and other forms expressed on flat surfaces), the plastic arts (sculpture, modeling), the decorative arts (enamelwork, furniture design, mosaic, etc.), the performing arts (theatre, dance, music), music (as composition), and architecture (often including interior design)."

% "Branches of the humanities include law, languages, philosophy, religion and mythology, international relations, gender and women’s studies, multicultural and regional studies, popular culture, and art and music, while branches of the social sciences include sociology, anthropology, archeology, geography, political science (including politics and government), psychology, communication studies, criminal justice, demographics, library and information science, and economics."

Arts and the humanities have long been reflections of human experience, emotions, and philosophical introspection~\cite{pool2018looking}. These domains, deeply rooted in subjectivity, creativity, and a nuanced appreciation of the world, have served as repositories of our history, culture, and identity. Over the past few years, however, the boundary between human creativity and machine computation has started to blur, ushering in an era where Artificial Intelligence (AI) influences artistic creation and reshapes our understanding of humanities.

Historically, AI's foray into domains requiring creativity was met with skepticism~\cite{ambartsoumean2023ai}. Critics posited that machines, bound by algorithms and devoid of emotions, could never truly comprehend or replicate the intricacies of artistic expression. Creativity was, after all, seen as the antithesis of computation, fueled by irregularities, out-of-box thinking, and a delicate understanding of the human condition. These very attributes, which are the cornerstones of arts and humanities, seemed out of reach for artificial entities.

\begin{figure}
  \begin{center}    \includegraphics[width=1\textwidth]{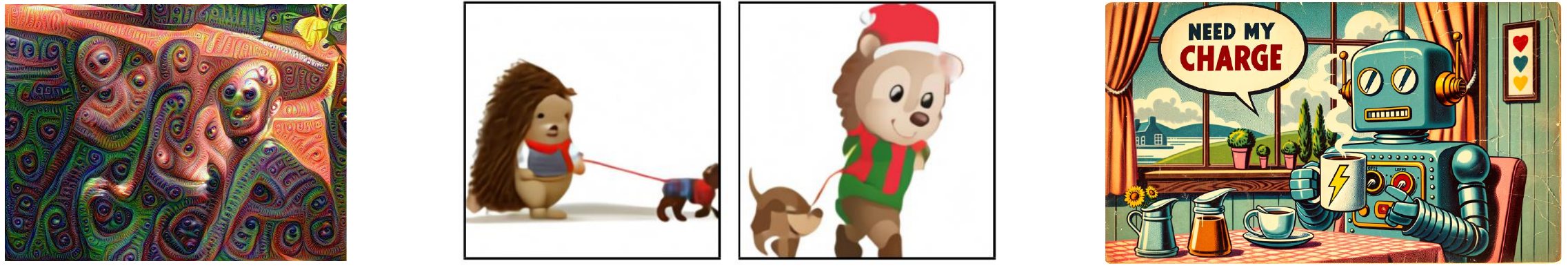}
  \end{center}
  \vspace{0cm}
  \caption{Some examples of AGI-generated images. \textbf{Left}: A heavily deep-dream-style photograph expressing "three men in a pool", which is difficult for humans to understand. 
  \textbf{Middle}: An image generated by DALL-E through translation from "an illustration of a baby hedgehog in a christmas sweater walking a dog"~\cite{ramesh2021zero}. 
  \textbf{Right}: Image created by DALL-E 3 with the prompt "vintage 1940s cartoon featuring a robot holding a steaming coffee mug with a lightning bolt symbol on it, text bubble that reads `Need my charge', sitting at a table by bay window in a coffee shop interior". The model can generate the high-quality image, and correctly understand the instruction.
  }\label{fig:overall}
  \vspace{0cm}
\end{figure}

More recently, the landscape has begun to shift. Early algorithms such as "Deep Dream" (2015) ~\cite{mordvintsev2015deepdream} and various approaches in the theme of "Neural Style Transfer"~\cite{singh2021neural} marked AI's early attempts at artistic endeavors. 
However, Deep Dream was plagued by the problem of generating repetitive canine facial features within images, and the style transfer process, while artistically intriguing, lacked the ability to create entirely new content or comprehend the underlying semantics of images.
% However, we have also witnessed that people's criticisms of the value of artificial intelligence in artistic creation are steadily softening. 
Nevertheless, these earlier attempts began a noticeable shift in perceptions, with increasing acceptance of artificial intelligence's contribution to artistic endeavors. 
A pivotal moment highlighting this change was when ``Edmond de Belamy, from La Famille de Belamy'', a portrait produced by Generative Adversarial Networks (GANs) \cite{goodfellow2020gan}, was sold by Christie’s New York on Oct 19, 2018 for \$432,500 \cite{cohn2018nyt}, which is more than 40 times Christie’s initial estimate. Despite facing skepticism and questions regarding its originality from other artists who work with AI, 
these rudimentary techniques marked the nascent stages of AI-assisted artistry. 

% While their outputs were novel, they were also a testament to AI's growing capability to engage with artistic processes. 

The leap forward came in 2021 with the arrival of text-to-image algorithms. 
Specifically, the introduction of DALL-E~\cite{ramesh2021zero}, supplemented by the unveiling of open-source projects like VQGAN+CLIP~\cite{esser2021taming,radford2021learning}, catalyzed the proliferation of AI art generators. 
Furthermore, in 2022, the release of "Stable Diffusion"~\cite{rombach2022high} by Stability AI and ``Imagen''~\cite{saharia2022imagen} by Google AI ushered in a new era of advanced AI-powered creativity. This release further democratized the Artificial Intelligence-Generated Content (AIGC) process.
The field of AIGC is still extremely young. Major contributors and platforms have a relatively short operational history, spanning less than a year. However, the trajectory suggests an impending turning point where AI capabilities will become sophisticated enough to revolutionize various art-related domains. For instance, in the realm of video game development, concept and traditional artists are already harnessing AI image generation for inspiration and as tangible assets in their creative works\footnote{\url{https://www.scenario.com/}}.
Looking ahead, once the complexities of image generation are comprehensively addressed, it is plausible that the intellectual capital steering this innovation will gravitate toward other modalities. This may encompass domains like auditory processing and generation, video synthesis, and literary generation, among other multidisciplinary challenges.

% The landscape, however, witnessed a paradigm shift in early 2021 with the advent of text-to-image algorithms. The introduction of DALL-E~\cite{ramesh2021zero}, supplemented by the unveiling of open-source projects like VQGAN+CLIP~\cite{esser2021taming,radford2021learning}, catalyzed the proliferation of AI art generators. Yet, by the culmination of 2021, only a modest increase in the number of such applications was observed.

% The year 2022 marked a significant inflection point. Momentum in this domain surged, culminating in the momentous release of "Stable Diffusion"~\cite{rombach2022high} by Stability AI, which further democratized the AI image generation process.
% 

With the recent advancement of Large Language Models (LLMs), the rise of Artificial General Intelligence (AGI) further challenges traditional perspectives. AGI~\cite{zhao2023brain}, with its potential to emulate holistic human cognition, promises not just to create art but to understand and appreciate it--and in fact many proponents of LLMs having early AGI capabilities espouse that these models already have a degree of understanding of the physical world and humans [see references. They need to be added to citations]. LLMs' integration into arts and humanities could revolutionize everything from literary synthesis, capturing the depth of human emotion, to creating multi-sensory art experiences and reinterpreting historical narratives.

This paper delves deep into the rapidly evolving nexus of AGI, arts, and the humanities. While celebrating the transformative potential of AGI, it also critically examines the following underlying questions: Can AGI truly be creative? Will it ever appreciate art the way humans do? And most importantly, as AGI blurs the lines between machine capability and human creativity, what does it mean for the future of arts and humanities? Through this discourse, we seek to navigate the promising yet perplexing frontier of AGI-infused artistry.
% \begin{figure}
%     \centering
%     \includegraphics[width=0.75\textwidth]{figures/main.png}
%     \caption{Potential risks of ChatGPT in healthcare.}
%     \label{fig:main}
% \end{figure}

% \textbf{Related surveys:}
% ~\cite{cao2023comprehensive, wu2023ai}

\section{Background}

\subsection{Generative AI: From GAN to ChatGPT}
A recent survey paper ~\cite{cao2023comprehensive} provides a comprehensive review of the field of AI-generated content (AIGC). AIGC refers to content like text, images, music, and code that is generated by AI systems rather than created directly by humans. 

The authors review the history of generative AI models, beginning with early statistical models like Hidden Markov Models and Gaussian Mixture Models. They then discuss the rise of deep learning models like GANs, VAEs, and diffusion models, with the transformer architecture (2017) identified as a key breakthrough that enabled large-scale pre-trained models like GPT-3 \cite{brown2020language}, ChatGPT \cite{liu2023summary}, and GPT-4 \cite{openai2023gpt}.

The paper categorizes generative models as either unimodal, which generate content in a single modality like text or images, or multimodal, which combine multiple modalities. For unimodal models, they provide an in-depth review of state-of-the-art generative language models like GPT-3 \cite{brown2020language}, BART \cite{lewis2019bart}, T5 \cite{raffel2020exploring} and vision models such as Stable Diffusion \cite{rombach2022high} and DALL-E 2 \cite{ramesh2022hierarchical}. 

For the multimodal generation, the survey examines vision-language models like DALL-E and GLIDE \cite{nichol2021glide} as well as text-audio, text-graph, and text-code models. These allow cross-modal generation between modalities. The authors discuss applications like chatbots, art creation, music composition, and code generation.

They also cover techniques that help align model outputs with human preferences, such as reinforcement learning from human feedback as used in ChatGPT. The paper analyzes challenges around efficiency, trustworthiness, and responsible use of large AIGC models. Finally, open problems and future research directions are explored.

\subsection{Opportunities and Challenges of General AIGC}

The enthusiastic reception of conversational agents like ChatGPT underscores AIGC's vast potential. However, researchers must grapple with critical challenges around data bias, computational efficiency, output quality, and ethical implications as AIGC rapidly gains traction \cite{wu2023ai}.

On the opportunities front, AIGC holds promise for boosting productivity in creative fields by acting as an intelligent assistant that can synthesize draft content. Cross-modal generation techniques can potentially bridge content formats, enabling applications like generating videos from text descriptions. In industry verticals like e-commerce, AIGC can scale the creation of catalog descriptions and customized landing pages. For news and entertainment, AIGC may enhance automation in production pipelines. The multi-task learning abilities of foundation models could spur innovation if applied judiciously. On the consumer side, AIGC can deliver more personalized, interactive, and immersive experiences.

However, substantial challenges remain. Massive computational resources are needed to develop and deploy the latest AIGC models~\cite{chen2023large}, which may concentrate power in fewer hands. More crucially, the data used to train AIGC models inherits human biases \cite{liu3surviving} that are reflected in outputs. Curating high-quality datasets is an arduous task. While human-in-the-loop approaches may improve model alignments, transparency and accountability are still lacking. Safeguards against toxic outputs remain inadequate as interactions uncover harmful edge cases. For high-stakes domains like healthcare, the risks of errors loom large. Despite great enthusiasm, researchers should adopt a measured approach while addressing these concerns through technical and ethical diligence.

While AIGC represents an exciting frontier for AI research with immense potential upside, responsible development calls for holistic solutions encompassing data curation and hygiene, efficient systems, user feedback loops, and transparency. With care and consideration for societal impacts, AIGC could usher in an era where generative AI assists and augments human creativity rather than displacing it. This survey provides a timely overview of the state-of-the-art as of late 2023, and a roadmap to guide progress in this rapidly evolving domain.

\section{Text Analysis and Generation}

Text analysis and generation are crucial domains in natural language processing influencing myriad applications. At the core, text analysis delves into comprehending intricate patterns, meanings, and sentiments in textual data, whereas text generation aspires to craft human-like text based on certain criteria or prompts~\cite{liu2023summary}. 
With the advent of sophisticated model architectures, the boundaries of what machines can comprehend and produce have been ceaselessly expanded. 
This section delineates the technical advancements underpinning these capabilities, including seminal models like Transformers, and extends into their pragmatic applications across diverse sectors such as poetry, music, law, advertising, and governance.

\subsection{Technical Advances}
The Transformer \cite{vaswani2017attention} architecture has undoubtedly carved a pivotal role in the progression of natural language processing models. Introduced by Vaswani et al. \cite{vaswani2017attention}, 
the architecture abandoned recurrent layers, traditionally used for sequence data, in favor of attention mechanisms. The core concepts emanating from this architecture, including encoders, decoders, BERT, and autoregressive language models, have since dominated state-of-the-art results in various NLP tasks.

\subsubsection{Transformer Architectures}
At the heart of the Transformer model lies the pivotal \textit{self-attention mechanism}, which computes a weighted sum of all words in a sequence relative to each other. This empowers the model to capture the intricate relationships between words, regardless of their positions in the sequence. 
Unlike recurrent models such as RNNs~\cite{sherstinsky2020fundamentals} or LSTMs~\cite{staudemeyer2019understanding} which process sequences iteratively, Transformers handle the entire sequence in parallel. 
This approach, coupled with additional design elements like positional encodings~\cite{gehring2017convolutional} and residual connections~\cite{hou2007saliency}, empowers Transformers to deliver both efficiency and effectiveness, even when confronted with lengthy sequences.

% \subsubsection{Encoders and Decoders}
% In the Transformer framework, encoders process the input data (e.g., sentences) in machine translation. Each encoder layer primarily consists of a multi-head self-attention mechanism and a subsequent feed-forward neural network, interleaved with normalization steps and residual connections. Decoders, on the other hand, are integral to tasks that necessitate output sequences. Their architecture mirrors that of encoders but with an additional multi-head attention layer that attends to the encoder's output. The decoder's final output undergoes a transformation through a linear layer followed by a softmax layer to generate predictions such as word tokens.

\subsubsection{BERT (Bidirectional Encoder Representations from Transformers)}
Emerging from the Transformer paradigm, BERT~\cite{devlin2018bert} represents a monumental shift in pre-training methods. Introduced by Google in 2018, this model captures bidirectional contexts by considering both preceding and following words in all its layers. BERT's pre-training phase involves a \textit{masked language model} objective wherein it attempts to predict randomly masked words in a sentence. Once pre-trained on vast corpora such as Wikipedia, BERT can be adeptly fine-tuned on specific tasks using small labeled datasets, by just adding appropriate task-specific layers~\cite{meyerson2017beyond}. 
This approach has made BERT highly versatile, allowing it to be applied to a wide range of NLP tasks.

\subsubsection{Autoregressive Language Models}
\textit{Autoregressive modeling}~\cite{brown2020language} uses a step-by-step approach, where predicting the next item in a sequence depends on what came before it. In the context of language modeling, when given part of a sentence, these models try to guess what words come next. 
Once properly trained, the models good at guessing what word comes after the previous ones in the sequence.
When autoregressive models generate text, they can use different methods, like beam search~\cite{freitag2017beam}, greedy decoding~\cite{gu2017trainable}, or probabilistic sampling~\cite{yu2022scaling}. 
A well-known example of an autoregressive language model is OpenAI's GPT (Generative Pre-trained Transformer)~\cite{radford2018improving}, which, in contrast to BERT's bidirectionality, is unidirectional and is primed mainly for text generation.

% \begin{figure}
% \centering
% % Use the relevant command to insert your figure file.
% % For example, with the graphicx package use
%   \includegraphics[width=14cm]{figures/prompt1.png}
% % figure caption is below the figure
% \caption{Some prompt examples, part one. (Image generated by DALL-E2)} 
% \label{fig:prompt1}       % Give a unique label
% \end{figure}
% \begin{figure}
% \centering
% % Use the relevant command to insert your figure file.
% % For example, with the graphicx package use
%   \includegraphics[width=14cm]{figures/prompt2.png}
% % figure caption is below the figure
% \caption{Some prompt examples, part two. (Image generated by DALL-E2)} 
% \label{fig:prompt2}       % Give a unique label
% \end{figure}

\subsection{Real-world Applications}
This section elucidates the diverse arts and humanities landscape of AGI by categorizing its applications into four distinct but interrelated subsections: 1) Literature Search and Analytics; 2) Linguistics and Communication; 3) Creative Endeavors. 
Each of these subsections represents a unique facet of AI's ever-expanding repertoire, showcasing its adaptability to perform tasks ranging from the systematic retrieval of knowledge to nuanced linguistic interactions, from meticulous analytics to imaginative, creative endeavors.

\subsubsection{Literature Search and Analytics}
Large language models, once equipped with common sense knowledge, can provide valuable assistance in various liberal arts domains such as history, classics, and philosophy that heavily rely on literature search and analysis. Based on~\cite{kaddour2023challenges}, LLMs can be helpful in the following aspects.
\begin{itemize}
    \item \textbf{Automated Literature Review:} LLMs can quickly scan and summarize large volumes of text. They can identify key concepts, themes, and relevant passages, saving researchers significant time and effort. 
    \item \textbf{Cross-Referencing:} LLMs can cross-reference texts, identifying connections and references between historical events, philosophical works, and classic literature, helping researchers explore intertextual relationships.
    \item \textbf{Summarization:} LLMs can generate concise summaries of lengthy texts, making complex philosophical or historical writings more accessible to a broader audience. 
    \item \textbf{Question Answering:} LLMs excel in answering specific questions related to historical events, philosophical theories, or classic literature. They can provide concise and accurate responses by drawing on their vast knowledge base.
    \item \textbf{Content Generation:} LLMs can assist in generating preliminary content by retrieving relevant information in the literature~\cite{li2022survey}. They can provide background information, context, and even propose arguments based on the input provided.
    \item \textbf{Teaching and Learning:} LLMs can be used as educational tools to provide explanations, generate practice questions, and engage students in discussions related to historical, classic, or philosophical topics.
\end{itemize}
Some specific examples of the ongoing and potential applications are provided as below.

\begin{figure}
  \begin{center}    \includegraphics[width=1\textwidth]{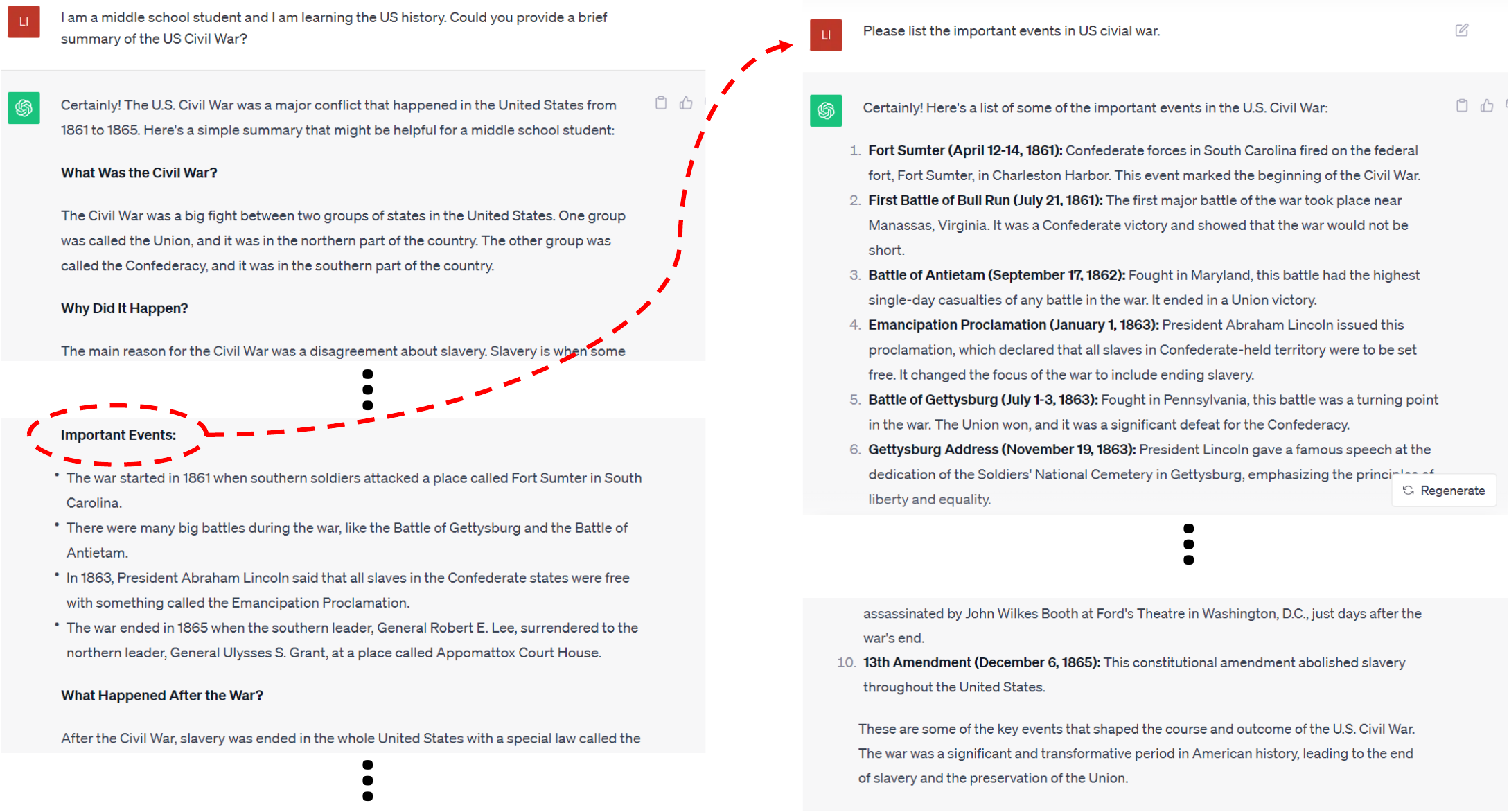}
  \end{center}
  \vspace{0cm}
  \caption{An example of using GPT-3.5 for learning history. The right part shows a follow-up question regarding the answer of the first question in the left part.
  }\label{fig:text}
  \vspace{0cm}
\end{figure} 

\textbf{Anthropology:} 
LLMs can process large amounts of anthropological data. 
% , including ethnography, archaeology, and cultural anthropology data, extracting information, patterns, and trends
Through text mining, LLMs deeply research documents, interviews, and historical texts to find information on specific cultures, social groups, or topics, and promote a deeper understanding of human social evolution and cultural differences~\cite{liu2023summary}. 
Second, LLMs can help analyze social surveys and public opinion polls, to gain an in-depth understanding of attitudes, beliefs, and behaviors in human society, helping researchers understand social trends and changes in public opinion~\cite{zhao2023survey}. 
Finally, through cross-cultural research, LLMs support the comparison of similarities and differences between different cultures, provide translation services, analyze cross-cultural communication, and conduct in-depth research on global issues such as globalization and cultural exchanges. 
%Possible prompt: "How have modern global migration patterns influenced cultural assimilation and the evolution of traditional practices in indigenous tribes?"

\textbf{Classics:} LLMs contribute to art historical research, providing in-depth insights into a specific period or style by analyzing textual descriptions of classical artworks, historical documents, and the lives of relevant artists. 
% Researchers can use LLMs to analyze artists and works and conduct in-depth research into the artist's life, creative process, and themes, techniques, and styles of artistic works, which will help to better understand the creative motivations and artistic development of classical artists. 
In terms of art education and popularization, LLMs can assist in the creation of art education materials, explain works of art so that more people can understand and appreciate classical art, and generate explanatory texts on art history for use in education, museum exhibitions, and cultural dissemination~\cite{kasneci2023chatgpt}. 
% Possible prompt: "Given the vastness of the Roman Empire and its influence on various cultures, how did local traditions in conquered regions blend with Roman practices to create unique hybrid cultures?"

% \textbf{Comparative Literature: } LLMs have many uses in the field of comparative literature such as analyzing different literary works to help researchers understand different literary traditions, styles, and themes; processing multilingual texts to reveal the linguistic differences and cultural characteristics of literary works; supporting cross-cultural comparative research and deepening understanding of literary works. Understand the diversity of global literature; identifying themes, symbols, and metaphors in literary works and providing deep insights~\cite{kandpal2023large}; providing tools for literary critical analysis to evaluate the literary value and style of works; processing literary historical documents to help trace the development of literature; analyzing language and style, to help identify the author's personalized writing style; generating explanatory texts, providing teaching materials and auxiliary materials for literary education, and promoting students' understanding and analysis of literary works~\cite{kasneci2023chatgpt}. 
% Possible prompt: "Analyzing literature from Eastern and Western traditions, how do hero archetypes differ, and what does this reveal about societal values?"

\textbf{Philosophy:} LLMs can be used for literature reviews to help researchers understand the current state of research on specific philosophical issues or thinkers~\cite{chang2023survey}. They can also analyze philosophical texts and understand the author's ideas, argument structure, and logic. 
% Scientists can use LLMs to generate new philosophical texts, explore different ideas and perspectives, and promote innovation and discussion in the field of philosophy. 
In addition, LLMs can also be used to analyze the structure and effectiveness of philosophical arguments, helping researchers better understand and evaluate philosophical papers. 
% Possible prompt: "Comparing Eastern and Western philosophical traditions, how do their perspectives on the nature of consciousness and reality converge and diverge?"

\textbf{Psychology:} LLMs can conduct literature reviews to help researchers understand the latest research and theories on specific psychological topics~\cite{wei2022emergent}. LLMs can also generate questionnaire questions to ensure they are clear and effective. In addition, LLMs can analyze comments on social media to understand emotional health and mental health issues of people, and provide support and resources. 
%Possible prompt: "How do cultural norms and societal structures influence individual cognitive development and perception of self across different societies?"

\textbf{History:} LLMs can analyze historical texts to help understand events, generate summaries, extract key information, and improve information processing efficiency~\cite{chen2021evaluating}. 
LLMs can also calibrate time in historical text, track events, and help establish a detailed historical timeline. They can also help extract character relationships, help build a relationship map, and conduct in-depth research on the influence of historical figures. 
% We can also use LLMs to conduct a literature review and quickly understand the current research status of a specific historical period, process digital historical files, and help build a more comprehensive historical database. 
% Possible prompt: "How did the Silk Road not only facilitate trade but also lead to the interchange of ideas, religions, and technologies between the East and West?"

\begin{figure}
  \begin{center}    \includegraphics[width=1\textwidth]{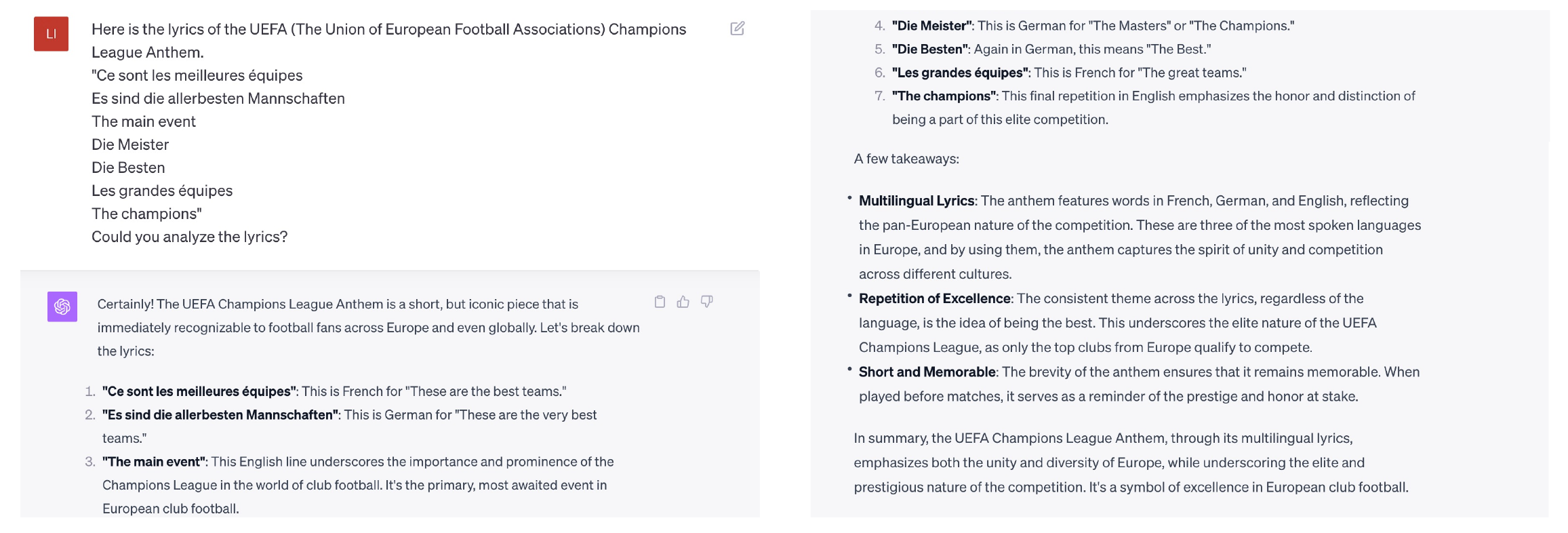}
  \end{center}
  \vspace{0cm}
  \caption{An example of using GPT-4 to analyze the background and design philosophy of the lyrics of the UEFA (The Union of European Football Associations) Champions League Anthem. The AI model can easily handle the multilingual content, and even point out the "spirit of unity" and "diversity" behind the design. 
  }\label{fig:linguistics}
  \vspace{0cm}
\end{figure}

\subsubsection{Linguistics and Communication}

LLMs can be highly beneficial in applications related to linguistics and communications due to their natural language processing capabilities and extensive knowledge base. An example that shows some these abilities is in Figure~\ref{fig:linguistics}.
\begin{itemize}
    \item \textbf{Language Understanding:} LLMs can be used to analyze the structure, grammar, and semantics of languages, aiding linguists in their research on syntax, morphology, and linguistic phenomena.
    % \item \textbf{Language Learning:} LLMs can provide language learners with real-time feedback on pronunciation, grammar, and vocabulary usage. They can also generate practice exercises and quizzes.
    \item \textbf{Translation Assistance:} LLMs can assist linguists and translators in translating text between different languages, helping bridge linguistic and cultural gaps.
    \item \textbf{Sentiment Analysis:} LLMs can perform sentiment analysis on text, enabling businesses and organizations to gauge public sentiment towards their products, services, or policies.
    \item \textbf{Speech Recognition:} LLMs can enhance speech recognition systems, improving the accuracy of voice-to-text transcriptions.
\end{itemize}
Based on the above capabilities, some specific examples of the ongoing and potential applications of LLMs are provided as below.

\textbf{Linguistics:} LLMs can generate new language texts and expand understanding of grammatical structures and vocabulary usage~\cite{linzen2019can,baroni2022proper}. Scientists can conduct semantic analysis through LLMs and conduct in-depth studies of lexical meanings, contextual relevance~\cite{linzen2021syntactic}, and semantic relationships of language expressions~\cite{chang2023survey}. These models can also be used to develop language learning tools to help students learn vocabulary, grammar rules, and other language knowledge. In studying language disorders, LLMs can reveal the manifestations and effects of language disorders in different contexts. However, some scholars have raised objections, believing that LLMs lack human cognition~\cite{katzir2023large}. 
% Possible prompt: "Phonological shifts in languages over time can be indicative of broader cultural changes. How have these shifts in major languages mirrored historical events?"

\textbf{Language Studies:} LLMs can analyze the grammatical, semantic, and pragmatic features of different languages. Based on this, LLMs can generate teaching materials and exercises with explanations to provide students with strong support in learning grammar, vocabulary, and expressions. 
In addition, LLMs excel in translation, supporting cross-language communication and translation \cite{devlin2018bert}. 
At the same time, LLMs play an important role in writing and creation and can create articles, compose essays, and generate various literary works. 
% By analyzing the language style in texts, LLMs can also help researchers understand the language characteristics of different authors, periods, or text types. 
In addition, LLMs support speech recognition technology, which allows speech input to be easily converted into text, facilitating speech interaction and speech recognition applications~\cite{kasneci2023chatgpt}. 
% They can also identify emotional colors in texts, helping to deeply understand and analyze emotional factors~\cite{zhang2023sentiment} in English texts. 
By processing large amounts of historical texts, LLMs can also assist researchers in tracing the evolution, change, and development of the language. 
% Overall, LLMs play a key role in building various natural language processing (NLP) applications, including question answering systems~\cite{mai2021geographic,ouyang2022instructgpt,openai2023gpt}, language understanding, and information retrieval~\cite{yasunaga2022deep}. 
% Possible prompt: "How has the evolution of the English language, particularly the influx of loanwords, reflected the socio-political dynamics of different eras?"

\textbf{Communitcation Studies:} LLMs can analyze large amounts of news and advertising, to reveal patterns, trends, and factors that influence the spread of information. 
LLMs can also analyze emotions and interactions on social media~\cite{zhang2023sentiment}, studying how information spreads in social networks and its impact on public opinion~\cite{dai2023ad}. 
% LLMs also analyze emotions in texts \cite{zhang2023sentiment}, helping scientists understand the emotional factors in information dissemination, and are of great value to the study of public opinion and emotional communication.
Researchers use LLMs to analyze the language and framing of news reports and study the way news media report events and their impact on audience perceptions. 
% LLMs help researchers analyze the structure of social networks and information dissemination networks, and gain an in-depth understanding of how information spreads in the network and the role of key nodes~\cite{kandpal2023large}. 
In terms of multilingual content, LLMs have the advantages of translation and understanding and are helpful in studying language differences and cultural factors in cross-cultural communication. 

\begin{figure}
  \begin{center}    \includegraphics[width=1\textwidth]{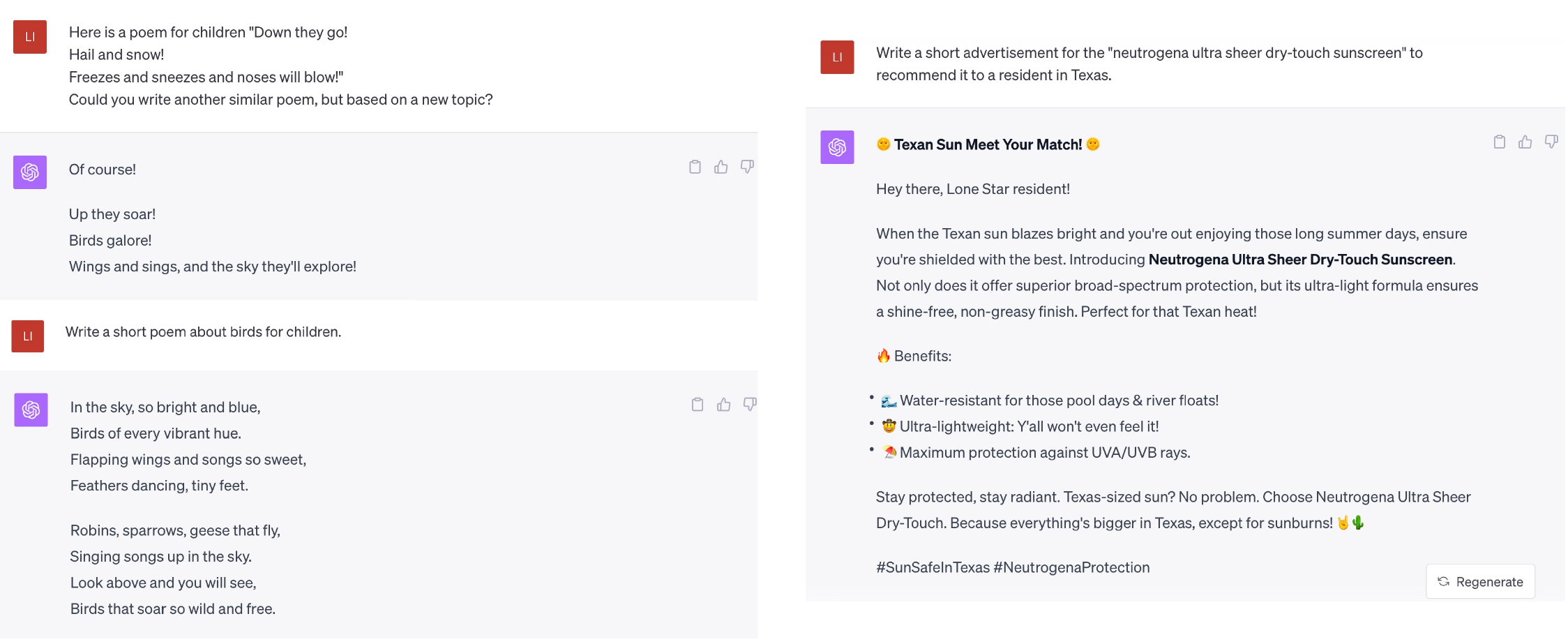}
  \end{center}
  \vspace{0cm}
  \caption{An example of using GPT-4 to write poems (left) and personalized advertisement (right). 
  }\label{fig:poem}
  \vspace{0cm}
\end{figure}

\subsubsection{Creative Endeavors}

This subsection starts with examples of several applications where AI might models generate "creative" contents, followed by a further discussion of whether AI can genuinely attain a level of creativity comparable to that of humans.

\textbf{Song Lyrics: } LLMs such as the GPT family can can write song lyrics that "tell coherent stories with rhyming words"~\footnote{\url{https://towardsdatascience.com/writing-songs-with-gpt-4-part-1-lyrics-3728da678482}}. Based on that, the AI models could be used to create new melodies to accompany the lyrics. GPT-4 is significantly better than GPT-3.5 at this due to better reasoning, complex instruction understanding, and creativity.
% Music melodies with different melodies were shown to GPT-4 and GPT-3.5. Both large language models can well understand the changes in music melody and create music works of the same style based on this. 
% writing song with GPT-4~\footnote{\url{https://towardsdatascience.com/writing-songs-with-gpt-4-part-1-lyrics-3728da678482}}

\textbf{Poetry:} There have been some research work on intelligent poetry writing and intelligent couplets~\cite{he2012generating}. However, the continuous development of LLMs has greatly facilitated research in this area. Figure~\ref{fig:poem} shows an example of using GPT-4 to write poems. Besides, a website~\footnote{\url{https://writeme.ai/poetry/}} shows the procedure to write a poetry using LLMs with only four steps. 
% The first step is to select the output language. The second step is to enter the keywords, phrases, themes, etc. that the author wants the poem to contain. The third step is to select the writing tone, such as convincing, doubting, etc. Enter Main Idea in the last step. After such simple input, a poem is created.

\textbf{Advertising:} 
The creation of effective and creative advertisements is a collaborative process that engages professionals with diverse skills and roles. Nonetheless, it is conceivable that certain roles may be assumed by Large Language Models (LLMs) in the future.
LLMs can help advertisers and marketers in creating content faster and potentially with quality akin to that of human content creators (see Figure~\ref{fig:poem}).
Moreover, given the abundance of successful advertising case studies available for reference in the field, LLMs with strong transfer capabilities such as GPT-4 can further improve the accuracy of advertising word generation through multi-shots to achieve the results desired by users~\cite{zelch2023commercialized}. 
LLMs can also analyze the promotional trends across a broad spectrum of advertisements, which enables conducting more efficient research, gaining deeper understanding of customer preferences, and addressing the complexities tied to information summarization~\cite{rivas2023marketing}.
% How will generative AI pay for itself? Besides user subscriptions, selling advertising is an option.
% , including commercial products and election candidates
% UGA has a professor (Jooyoung Kim) working on this(Can not find the reference up till now) 
% thereby helping to summarize the current market tendencies and greatly reducing the difficulty of information extraction and summary

With the rapid development of AI models, a question arises: Will AI eventually replace human creativity, or will humans continue to be the paramount source of innovation and originality?
A brief creativity comparison between humans and AI is as follows. 
\begin{itemize}
    \item \textbf{Human creativity} is influenced by personal experiences, emotions, and imagination, while it has limitations in terms of time, resources, knowledge, and experience, in addition to external factors like societal and economic influences.
    \item \textbf{AI creativity} is primarily grounded in algorithms and data, so a dominant view is that AI can only work with previous data and patterns, and cannot come up with entirely novel ideas on its own.
    Moreover, AI's deficiency in emotion and empathy poses another restriction. It is unable to replicate human emotions or grasp the emotional depth that art or music carries, potentially resulting in AI-generated content lacking the profound emotional impact typically attributed to human creativity.
\end{itemize}

\section{Graphics Analysis and Generation}
% 2D images, 3D images, point clouds, designs

% static vs dynamic

% input type: image, text, ...

Graphics encompass various formats, including 2D images, 3D point clouds, 3D meshes, and design schematics. These can be categorized based on their nature as either static or dynamic. The input types for graphic generation and analysis can also vary, ranging from images, text, and even other multidimensional data sources.

\subsection{Technical Advances}
There are numerous technical advancements that have propelled the fields of graphics analysis and generation to new heights.

\subsubsection{Generative Adversarial Networks (GANs)} 

In the mid 2010's, GANs \cite{goodfellow2014generative} ushered in a new era in the field of image generation. At the heart of a GAN framework are two intertwined neural networks: the generator and the discriminator. 
The generator creates images either from random noise in the case of unconditional GANs \cite{goodfellow2020gan} or guided by text/categories for conditional GANs \cite{mirza2014cgan}. Concurrently, the discriminator evaluates these generated images against real images. Through iterative refinement and adversarial training within a minimax game framework, the generator refines its outputs, aiming to create images indistinguishable from real ones while the discriminator learns to be an increasingly better judge of real versus AI-created images. This adversarial process has led to the generation of exceptionally high-quality and realistic images, significantly surpassing previous methods such as autoregressive models, Variational Autoencoder \cite{pu2016vae}, and normalizing flows \cite{kobyzev2020normalizing}. Moreover, the versatility of the GAN framework has extended beyond traditional imagery modality to other graphics formats such as 2D/3D point clouds \cite{li2018point,shu20193d,ramasinghe2020spectral}, graphs \cite{wang2018graphgan}, 3D object shapes \cite{wu2016learning}, and so on.

\subsubsection{Style Transfer Techniques} 

Neural style transfer~\cite{jing2019neural} has emerged as a captivating application of deep learning in graphics. By leveraging the intricate structures within neural networks, style transfer algorithms can take the artistic style from one image and apply it to another, enabling the creation of unique, artistically rendered outputs.

\subsubsection{Generative Models for 2D Images} GANs excel in producing high-quality 2D images, often to the point of being indistinguishable from real photographs. Additionally, Variational Autoencoders (VAEs) offer a probabilistic framework to generate 2D images \cite{doersch2016tutorial} while capturing the underlying data distribution. Both models can utilize inputs such as noise vectors, existing images, or textual descriptions to guide the generation process. A seminal work in this domain is alignDRAW \cite{mansimov2015generating}, which generates captions for images based on VAE and an attention mechanism. 

\subsubsection{Generative Models for 3D Images and Point Clouds}  
GANs and VAEs have been extended to generate 3D voxel grids or point cloud representations \cite{li2018point,shu20193d,ramasinghe2020spectral}. Moreover, models like PointGAN \cite{li2018point} focus specifically on generating high-quality point cloud data, capturing intricate 3D structures. Inputs for these models can range from 2D projections, textual descriptions, or even other 3D structures for tasks like super-resolution in 3D space.

\subsubsection{Generative Models for Designs}
Design generation, especially for aspects like logos, user interfaces, or architectural layouts, has seen innovation through models like CreativeGAN \cite{nobari2021creativegan}. These models can take inputs in the form of design constraints, user preferences, or textual descriptions to generate design mockups. The produced designs can be static (like a logo) or dynamic (like an interactive UI prototype).

\subsubsection{Static vs. Dynamic Generation} \
While many generative models focus on producing static outputs, there's a growing interest in dynamic content generation, especially in domains like video synthesis or interactive designs. Recurrent neural networks (RNNs), especially the Long Short-Term Memory (LSTM) networks, combined with GANs (like VideoGAN \cite{vondrick2016generating}), as well as the recent video transformer \cite{arnab2021vivit,liu2022video} have made strides in generating video sequences. This aligns with the broader trend of moving from static images to dynamic, time-evolving sequences in synthetic media. We will discuss this in detail in Section \ref{sec:video_generation_tech}. 

\subsubsection{Diverse Input Types} 
A hallmark of modern generative models is their ability to handle a variety of input types. While noise vectors remain a staple, there is a growing trend of models using textual descriptions to guide synthesis, allowing for more controlled and descriptive generation. This has been evident in models like AttnGAN \cite{xu2018attngan} and Df-GAN \cite{tao2022df}, where textual descriptions can guide the fine details of image synthesis, ensuring alignment between described content and the generated image.

\subsubsection{Diffusion Models}

% Diffusion Models (DMs) \cite{croitoru2023diffusion} are innovative techniques that create images by sequentially applying a series of processes that "clean up" or "denoise" the image at each step. This unique approach has made them one of the best at synthesizing images and more. One great feature of these models is that they can be directed or controlled in how they generate images without the need for extensive retraining.

% However, there is a drawback: because DMs generally work directly with the individual pixels of an image, they require a tremendous amount of computational power and time. In fact, to optimize these models to their best performance can take hundreds of days using powerful graphics processing units (GPUs), and using them can also be costly in terms of resources.

% To make these models more efficient without sacrificing their performance, researchers have started training DMs using the underlying structures or "latent spaces" of already trained models, known as autoencoders \cite{kingma2019introduction}. This approach reduces the computational burden of the process while still retaining the important details that make the images look realistic.

Diffusion Models (DMs) \cite{sohl2015deep,ho2020denoising,nichol2021improved,song2020score} are innovative techniques conceptually inspired by non-equilibrium thermodynamics \cite{sohl2015deep}. These models progressively introduce Gaussian noise during the forward (diffusion) process and subsequently learn to reverse the diffusion process to reconstruct the image from noise by predicting the previously added noise and then denoising. This unique approach has made them one of the best at synthesizing images and more. One great feature of these models is that they can be directed or controlled in how they generate images without the need for extensive retraining.

\textbf{Denoising Diffusion Probabilistic Models:} However, the Denoising Diffusion Probabilistic Models (DDPMs) \cite{ho2020denoising}, as one of the pioneering works in diffusion models, have a drawback -- given the fact that both the diffusion forward process and the denoising reverse process in DDPMs involve long Markov chains which consist of thousands of steps and DDPMs generally work directly with the individual pixels of an image, they usually require a tremendous amount of computational power and time for both model training and image sampling. In fact, optimizing these models to their best performance can take hundreds of days using powerful graphics processing units (GPUs), and using them can also be costly in terms of resources.

\textbf{Denoising Diffusion Implicit Models:} Consequently, to tackle the low sampling speed issue, Denoising Diffusion Implicit Models (DDIMs) \cite{song2020denoising} was proposed as fast sampling diffusion models closely related to DDPMs. DDIMs maintain the same marginal noise distributions as DDPM but diverge with a non-Markovian diffusion process and deterministically map noise to images. Consequently, DDIMs can generate high-quality images while significantly reducing the generation steps from 1000 in DDPM to just 50.

\textbf{Conditional Diffusion Models:} In addition to the aforementioned unconditional diffusion models, researchers have developed DMs that are conditioned on additional inputs such as class labels, reference images, or text sequences \cite{dhariwal2021diffusion,ho2022classifier,saharia2022image,rombach2022high} to better guide the generation process. 

\textbf{Latent Diffusion Models:} To make DMs more efficient without sacrificing their performance, researchers have also started training DMs by using the underlying structures or ``latent spaces'' of already trained models, known as autoencoders \cite{kingma2019introduction}. This approach reduces the computational burden of the process while still retaining the important details that make the images look realistic.

\textbf{Stable Diffusion: }
Latent Diffusion Models (LDMs) \cite{rombach2022high} have been instrumental in advancing the domain of image synthesis. These models incorporate the robust synthesis capabilities inherent to traditional DMs but with an added advantage: the flexibility of operating in latent space. This transition to latent space doesn't just add flexibility, it also introduces a remarkable equilibrium. The LDMs are designed to minimize model complexity without compromising the richness of image details. As a result, there is a noteworthy improvement in visual fidelity in these models versus pixel-based DMs, making output images sharper and more true-to-life.
One of the standout features introduced to LDMs is the integration of cross-attention layers. This inclusion is not merely a technical enhancement but a transformation in adaptability. With these layers, LDMs are equipped to handle a diverse range of conditioning inputs. Whether it is textual data or bounding boxes, the model processes them with equal proficiency. This versatility is pivotal, especially when high-resolution image synthesis is the goal. LDMs have shown the capability to generate these detailed images using a convolutional approach, offering a blend of clarity and detail that was until recently very challenging to achieve.
% The advantages of LDMs are not restricted to their synthesis capabilities. 
Another advantage of LDMs is the low computational overhead. 
One of the pressing challenges in image modeling has always been the computational demands, especially with pixel-based DMs that tend to be resource-intensive. LDMs present a solution to this long-standing problem. Despite their advanced features and superior performance, they operate with a significantly reduced computational overhead. This efficiency ensures that high-quality image synthesis is not just the domain of those with vast computational resources but is accessible to a broader spectrum of researchers and practitioners using small GPU clusters or even desktop or mobile devices.

\begin{figure}
  \begin{center}    \includegraphics[width=0.9\textwidth]{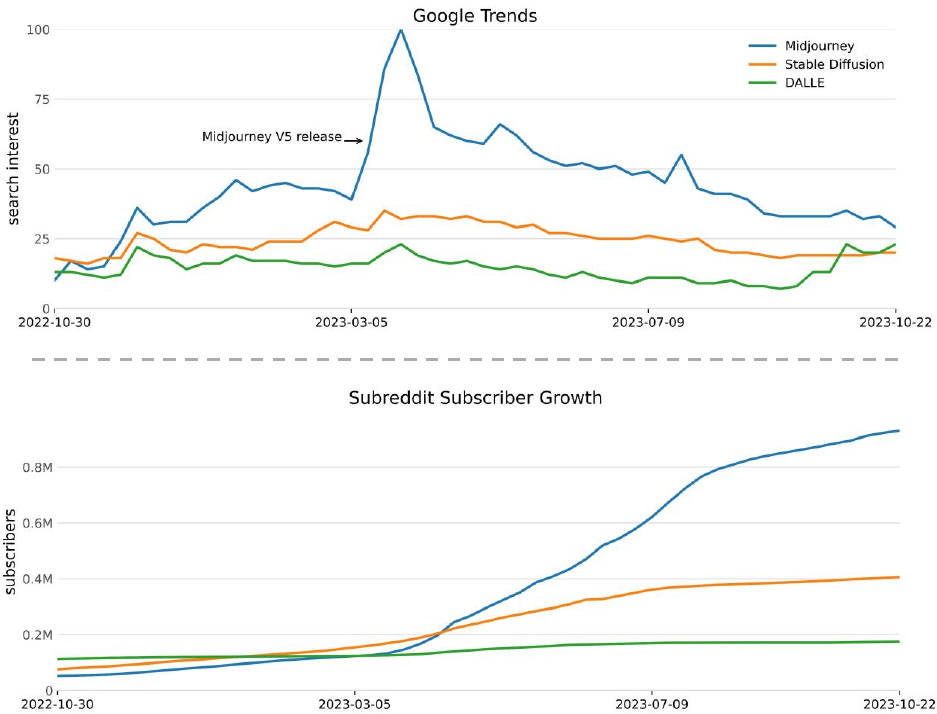}
  \end{center}
  \vspace{0cm}
  \caption{Google trends (top) and subreddit subscriber growth (bottom) for the past 12 months of the top 3 AI art generation tools: Midjourney, Stable Diffusion, and DALL-E. Data source: \href{https://trends.google.com/trends/}{Google Trends}  and \href{https://subredditstats.com/}{Subredditstats}.
  }\label{fig:diffusion_growth}
  \vspace{0cm}
\end{figure} 

\subsection{Real-world Applications}
% photos, paintings, websites, designs, ...
Recent advancements in generative AI, particularly in image models, have gained significant popularity not only in research but also in real-world applications. An increasing presence of AI-generated content (AIGC) can be observed in websites, advertisements, posters, and magazines. These models have the capability to generate diverse yet coherent graphics from cartoon illustrations to realistic photographs, eliciting interest across various industries. Figure \ref{fig:diffusion_growth} illustrates the trending popularity of renowned generative AI tools over the past year, indicating a promising future for their real-world applications.

\subsubsection{Cartography and Mapping} 
\begin{figure}
  \begin{center}    \includegraphics[width=0.9\textwidth]{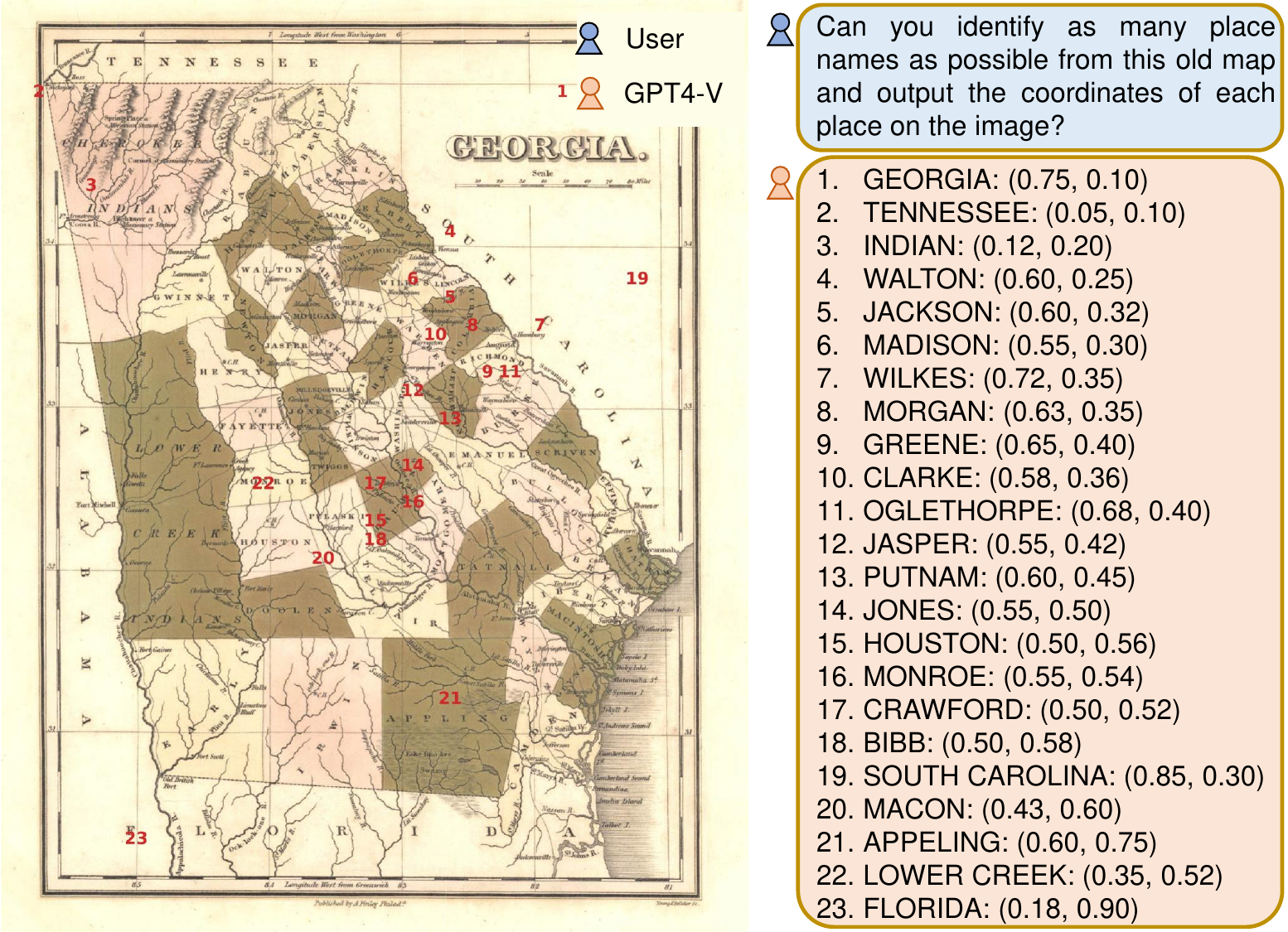}
  \end{center}
  \vspace{0cm}
  \caption{An illustration of using GPT-4V to do place name extraction and localization from a historical map of Georgia, USA as Kim et al. \cite{kim2023mapkurator} did. 
  The input to GPT-4V is the historical map and the prompt shown in the blue box. The answer from GPT-4V is shown in the orange box which provides a list of extracted place names as well as their map coordinates. Based on these map coordinates, we plot the corresponding numbers on the historical maps. 
  }\label{fig:map_ner}
  \vspace{0cm}
\end{figure}

As the field of studying, designing, and using maps, cartography is considered a discipline that encompasses both art and science by many cartographers \cite{krygier1995cartography}. Cartography includes various important scientific questions such as map projection \cite{mulcahy2001symbolization,chrisman2017calculating,mai2023sphere2vec}, map generalization \cite{brassel1988review,ai2001map,kang2020towards}, building pattern recognition \cite{he2018recognition,yan2019graph,mai2023towards},  drainage pattern classification \cite{yu2022recognition}, and so on. Because of the nature of cartography, most of these tasks require an AI model to manipulate or generate geospatial vector data (e.g., points, polylines, and polygons) \cite{mai2019space2vec,mai2023csp,mai2022review}. Although there are multiple existing foundation models, most of them are unable to handle this kind of vector data which makes these foundation models inapplicable for various cartography tasks \cite{mai2022towards,mai2023opportunities}. However, there are also various important cartography tasks that current foundation models are able to handle such as historical map data extraction. For example, various multimodal foundation models such as KOSMOS-2\cite{peng2023kosmos2} and GPT-4V \cite{openai2023gpt} can be used for extracting and linking text from a historical map of Georgia \cite{shbita2023building,kim2023mapkurator}. Figure \ref{fig:map_ner} shows one illustrative example and the response from GPT-4V. It's evident that even without task-specific fine-tuning, GPT-4V can identify various place names from maps. Additionally, although the accuracy may not always be very high, GPT-4V can generate map coordinates for these places. 
%Figure \ref{fig:land_survey_qa} illustrates another example in which we use GPT-4V to do tree-type prediction and localization from a land survey map made in 1832. We can see that GPT-4V can identify various tree types, but struggles to produce their corresponding coordinates. This challenge could arise from the low quality and contrast of the historical land survey map. However, the results generated in a zero-shot setting remain impressive. 
Moreover, foundation models can be also used for map reading and map-based question answering for topographic maps, thematic maps, or even narrative maps \cite{mai2022narrative}. Despite these success stories, applying foundation models and AIGC on cartographic applications can also lead to ethical issues such as inaccuracies, unanticipated features, and reproducibility \cite{kang2023ethics}. So the pros and cons of foundation models on cartographic applications need to be investigated further.

% \begin{figure}
%   \begin{center}    \includegraphics[width=0.9\textwidth]{figures/land_survey_qa.jpg}
%   \end{center}
%   \vspace{0cm}
%   \caption{An illustration of using GPT-4V to do tree type recognition and localization from a historical land survey map made in 1832. The input to GPT-4V is the historical land survey map and the prompt shown in the blue boxes whole the orange boxes depict the generated answers. 
%   }\label{fig:land_survey_qa}
%   \vspace{0cm}
% \end{figure} 

\subsubsection{Environmental Design}
\begin{figure}
\centering
  \includegraphics[width=15cm]{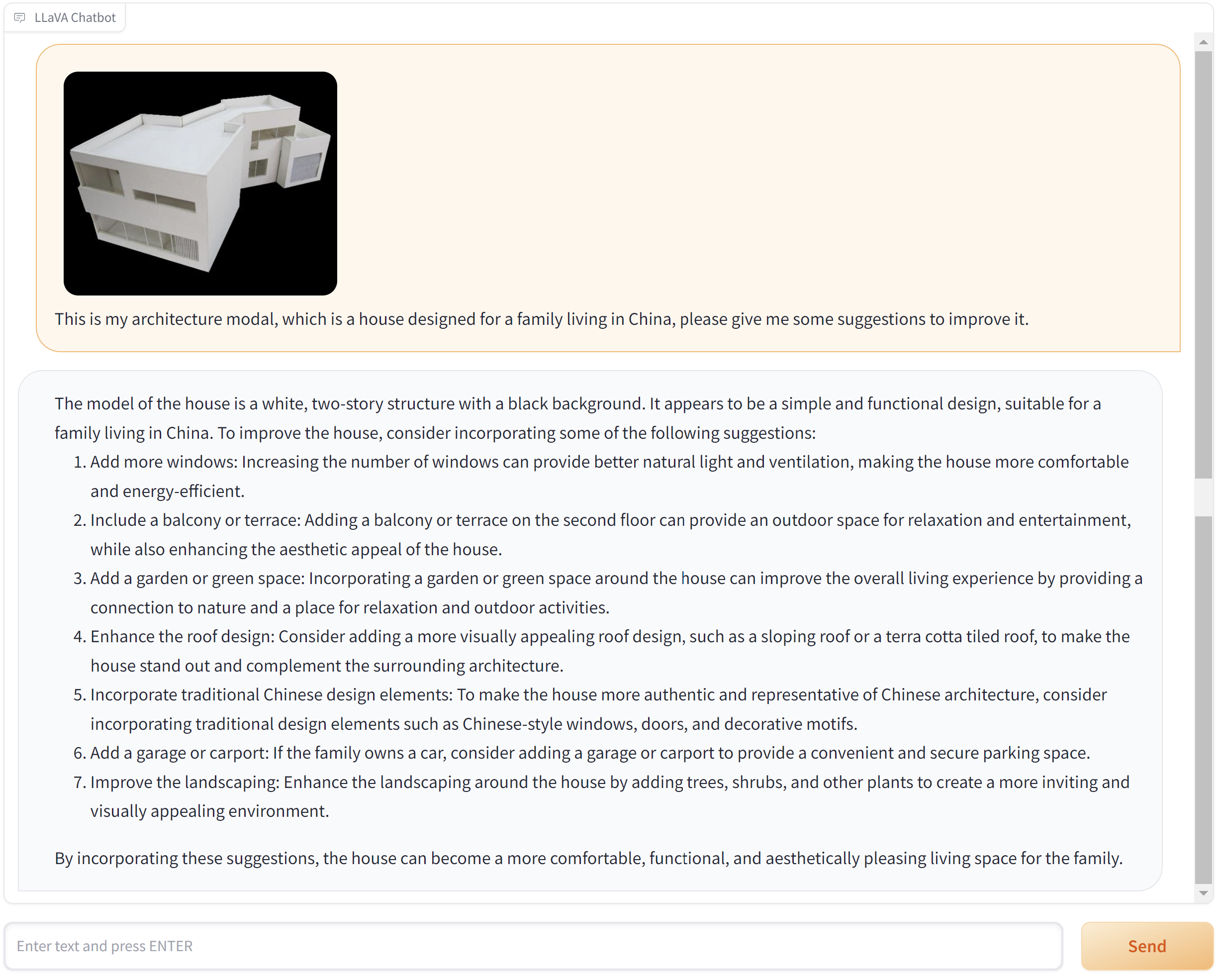}
\caption{An example of AI's suggestion for architecture design (Generated by LLaVA)} 
\label{fig:arch_design}       % Give a unique label
\end{figure}

AIGC, especially text-to-image generation, provides valuable tools for designers. These technologies can offer inspiration and improve workflow efficiency in the field of environmental design, including landscape architecture\cite{fernberg2023artificial,li2023designer}, urban design\cite{seneviratne2022dalle, sanchez2023prospects}, architecture\cite{ploennigs2023ai,chaillou2021ai}, and interior design\cite{he2023revamping,hussein2023improving}. In the initial design phase, AI sparks inspiration by generating diverse intentional images in various styles. It is particularly imaginative in the generation of special-shaped buildings\cite{perez2017blurring}. Providing diverse reference styles also helps to confirm the tone and style of the work. In the design review stage, AI can perform rapid partial replacement, helping designers to clarify the replacement effect and improve the speed of modification. Take architecture as an example\cite{stigsenai}. Designers can compare the effects of different surface materials, body proportions, and facade details with AI. In design analysis, AIGC’s ability to generate images with multiple perspectives and scales supports designers in producing analysis diagrams such as streamlining analysis and functional partitioning. Finally, after the design plan is finalized, AI can accelerate rendering and offer dimension choices, like spatial scale, weather, and night scenes. Since environmental design is a graph-oriented industry, the application of newly emerging multi-modal foundation models (FMs) in this field is in a more auxiliary position compared to text-to-image generative AI. Multi-modal FMs can assist designers in understanding statistic diagrams and then enhance scientific support for designs. They can also identify and illustrate images, including remote sensing images, architecture, and interior photos, which can be used for case studies and style reference. They can even evaluate design works and give suggestions for improvement. Figure \ref{fig:arch_design} shows an example of LLaVA’s recommendations for architecture design work. In this example, LLaVA extracts several building features like windows, balcony, garden, and roof from a photo of an architectural model, as well as information from the prompt to offer advice. This example proves LLaVA’s capacity to analyze architecture functionally, although it has not shown insights into aesthetic and social meanings, which remain the exclusive domain of architects.

\subsubsection{Photography and Editing}
\begin{figure}
\centering
  \includegraphics[width=14cm]{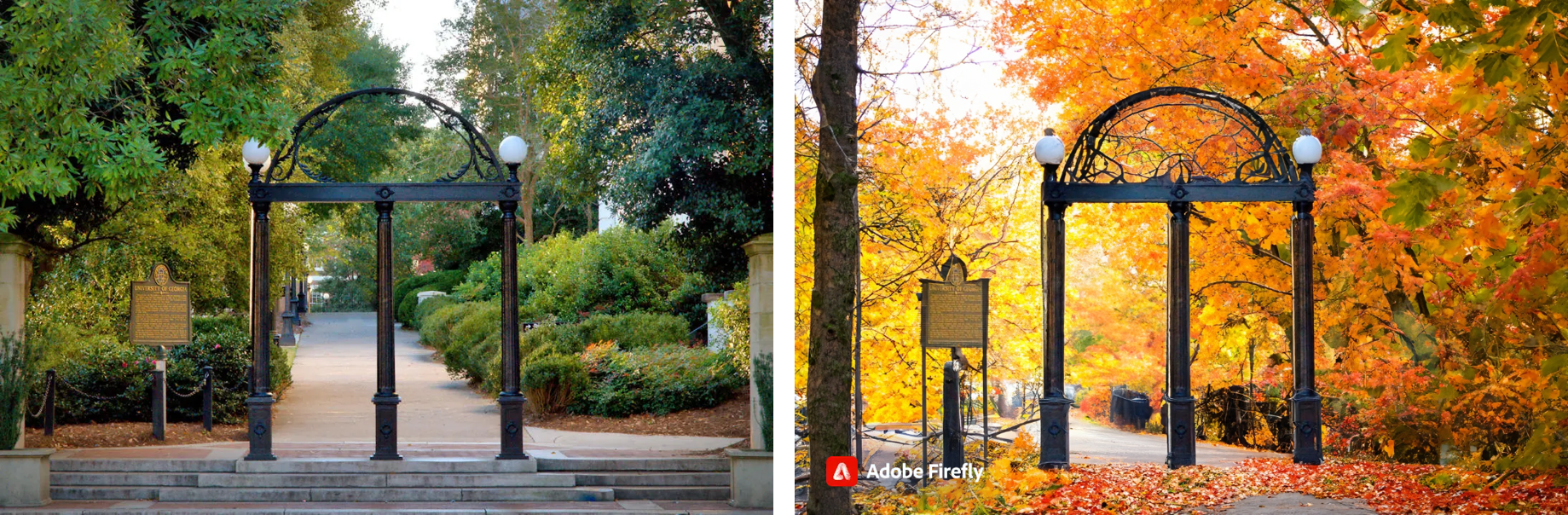}
\caption{An example of Photo editing. \textbf{Left}: Before editing. 
  \textbf{Right}: After editing by Adobe Firefly with the prompt "Turn the background into autumn".}
\label{fig:Adobe_phto_editing0}       % Give a unique label
\end{figure}

\begin{figure}
\vspace{8pt}
  \begin{center}    \includegraphics[width=0.9\textwidth]{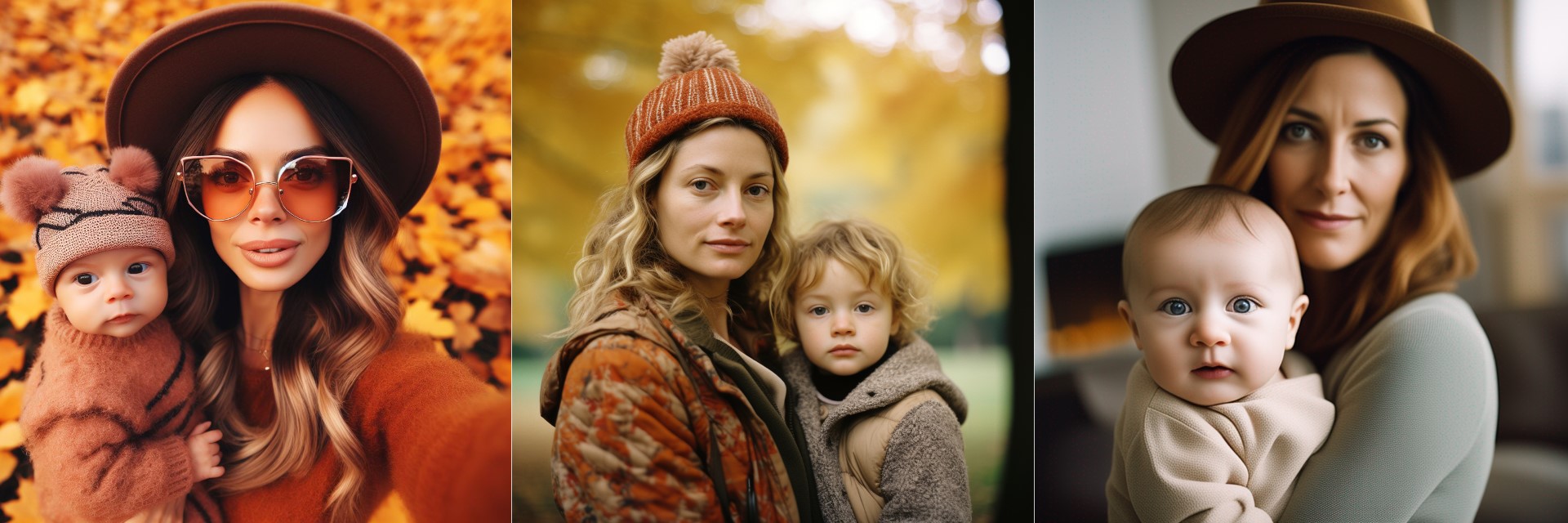}
  \end{center}
  \caption{Synthetic photography generated by three current generative diffusion models: DALL-E 3 (left), Midjourney (center), and Stable Diffusion XL (right). All images are generated by the following identical prompt for image generation: ``hip mother age 30 looking from the baby's perspective, lens: 35mm, focus: mother's face, style: modern realistic, fashion: chic, fall colors, no patterns.'' 
  }\label{fig:art_generation}
  \vspace{0cm}
\end{figure}

Foundation models and AIGC have transformative forces to improve image quality and image editing efficiency, and even extend the domain of photography to artificially generated ``photographs.'' In terms of image quality, AIGC can be used for refining and upscaling historical and/or low-resolutions photos with the so-called image restoration \cite{kawar2022ddrm,zhu2023denoising} and image superresolution \cite{saharia2022sr3,mai2023ssif} capability. Furthermore, AIGC’s capacity to eliminate reflections saves a lot of photos ruined by reflections from glasses. When it comes to enhancing photo editing efficiency, foundation models shine in various aspects. Firstly, FMs such as SAM \cite{kirillov2023sam} excel in objective segmentations without any model finetuning. 
%can easily identify objects like humans and animals, and cut out their outlines. 
Secondly, with the so-called image inpainting \cite{rombach2022high,kawar2022ddrm,song2022objectstitch} ability, FMs can be applied to remove the recognized objects and automatically replace the target area with a coherent background. 
Figure \ref{fig:Adobe_phto_editing0} shows one example with Adobe Firefly in which the background of University of Georgia's Arch is changed from summer to autumn style. 
FMs can be also used to generate the missing part of an object when this object is blocked or outside of the image scene. 
In addition, FMs can change the characteristics of these objects, including color, texture, and even style, which used to be a time-consuming task. 
% The most creative part of FMs in photo editing is their capacity to generate content tailored to photos\cite{song2022objectstitch}. 
% For instance, if the identified object is blocked, AI can generate the blocked part. 
FMs can also generate entirely new objects from text prompts or scribbles  \cite{singh2022paint2pix} which perfectly fits the light conditions and angle of the photo. Finally, Diffusion Models (DMs) are excellent at creating photorealistic images from text prompts or other images \cite{ho2022imagen}. This synthetic image generation is a step-change in creative photography, and even calls into question the traditional definition of photography, which has traditionally been associated with recording photons onto analogue or digital recording devices (e.g., chemicals on a plastic sheet or CMOS image sensors). Figure \ref{fig:art_generation} illustrates three synthetic photographies generated by three widely used generative diffusion models.

\subsubsection{Illustration}

While currently, it cannot entirely replace professional illustrators, AIGC significantly aids the initial conceptualization stage\cite{mustafa2023impact}, much like its role in environmental design. AIGC’s rapid iterations from inspiration to finished drawing allow illustrators to quickly determine the composition, elements, and style of a painting. Since AI has significantly reduced the difficulty of illustration, in situations where super-precise drawings are not needed, the images AIGC generates can even be directly applied to illustrated books, comics, and print advertisements. Illustrators use LLMs to give instruction for storyboards, character design, and painting style, then employ multi-modal FMs to generate complete illustrations\cite{lu2023could}, that have consistency in scenes, characters, and style. With the help of image editing tools, some of which are also powered by AIGC, typesetting work can also be completed. Using the tools mentioned above, AIGC completely supports the entire illustration process and effectively reshapes the traditional workflow.

\subsubsection{Graphic Design}
AIGC has a wealth of applications in graphic design, including logo design, print advertising, product packaging design, and mock-ups\cite{matthews2023destroy}. AI-generated logos can be suitable for printing or can be used in richer scenarios such as storefront signs and building facades after fine-tuning by the designer. The production process of print advertisements with AI is close to that of illustrations. These tools' fast, low-cost, and uniform style has made them favored tools among print advertisers. AIGC also has a role in product packaging design and mock-ups, where it can generate packaging for a series of products under the same subject, and provide a variety of usage scenarios for products. These processes replace traditional photography or rendering, greatly reducing time and cost expenditures. 

\subsubsection{Font Design}
\begin{figure}
\centering
  \includegraphics[width=1\textwidth]{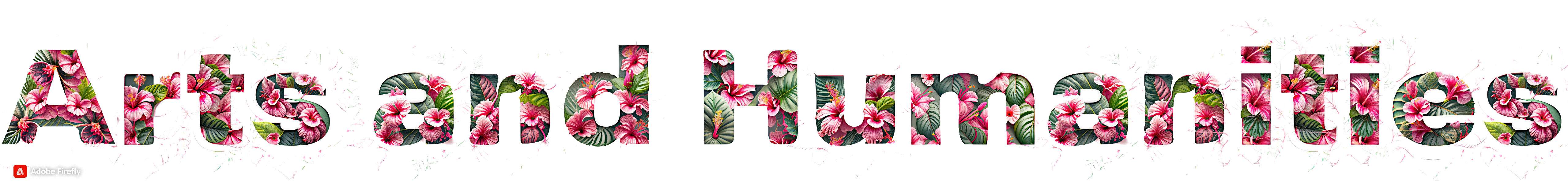}
\caption{An example of font design generated by Adobe Firefly with the prompt "pink hawaiian hibiscus flowers and leaves realistic, and the shapes of flowers and leaves can be out of letters" and the text "Arts and Humanities".}
\label{fig:Adobe_Arts and Humanities}       % Give a unique label
\end{figure}

AI provides a rapid and easy way to conduct font design. Supplying AI with letter references, whether in vector form or rough hand-drawn sketches, the AIGC models are capable of comprehending their unique style and seamlessly adapting it by maintaining harmonious counters and bodies. These refined letter forms can then be seamlessly integrated into existing texts\cite{yuan2021font}. In addition to learning variables in type design, AIGC also treats references as graphs, considering elements such as color, texture, shading, reflection, glow, or other effects\cite{yuan2020art}. Therefore, these graphic features can be transferred to letters. Moreover, natural language also provides enough information to design new fonts. Figure \ref{fig:Adobe_Arts and Humanities} is an example of art font design accomplished by Adobe Firefly. Both texture and shape are successfully generated according to the prompt, although there are some imperfections around the edges. %This is the process of using AI to create artistic fonts.

\subsubsection{3D Design}
3D design plays a pivotal role in the animation, video game, and film industries. The application of AIGC and FMs in this domain unfolds into two categories. One use case is text-to-3D, an extension of image generation\cite{zhang2023text2nerf}. Just as %AI generates images from text prompts, 
text-to-image generative models, 
FMs can be used to generate 3D models based on text prompts. This can be applied in the prototyping of scenes and characters, enriching the creative process. The other use case is 3D model manipulation. Given a 3D model, AI can adjust its posture automatically according to reference pictures or user instructions instead of manually adjusting joint positions\cite{yao2022learning}. This feature caters not only to professional designers but also fosters accessibility for novice users in 3D model creation. Moreover, the surface of 3D models can be also generated by text prompts. Combined with image generation models, AI-enhanced 3D models improve the efficiency of 3D character generation, scene rendering, and even product design.

\subsubsection{Fine Art} 
Perhaps most divisively, AI-based image generators can be used to create works traditionally associated with fine art\cite{chen3using}, or art with no purpose but to amaze and please its audience. Fine art painting and photography are considered the epitome of human skill and creativity, yet AIGC, specifically in the form of diffusion models (DMs), has created work that many consider on the level of highly skilled photographs and paintings. Needless to say, many practitioners and critics state that DMs cannot now or ever replace human creativity and skill. At present there is no clear answer to the question of whether AI, or AI in combination with a human, will be able to create work on the level of the highest human artistic achievements, but this is an area that should be watched in the coming months and years.

\subsubsection{Evolutionary Creativity}

Evolutionary art and evolutionary music are innovative fields of generative AI \cite{romero2008art}. They belong to the general field of evolutionary creativity and leverage Evolutionary Computation to generate esthetically pleasing visual arts or music. Evolutionary computation \cite{eiben2015introduction} is a collection of methods based on the principles of Darwinian Evolution. They simulate a population of solutions evolving over time through operations of selection, mutation, and recombination, better solutions are found. The field of evolutionary creativity includes multiple approaches which could be divided into human-in-the-loop approaches where the evolutionary algorithm generates art or music and a human either assigns a score (fitness) or compares different pieces of art or music and picks the best. Other approaches rely on an objective measure of merit based on some rules of thumb in music composition for example.

% \section{Audio Analysis and Generation}

% \subsection{Technical Advances}

% \subsection{Real-world Applications}

% \textbf{Music:} text to music~\footnote{\url{https://www.theverge.com/2023/1/28/23574573/google-musiclm-text-to-music-ai}}

\section{Video and Audio Analysis and Generation} \label{sec:video_generation}

\subsection{Technical Advances} \label{sec:video_generation_tech}

Video content (including audio) is a predominant form of information consumption and communication in the digital age. With the exponential growth in video data, there arises an acute need for effective video analysis and generation tools powered by artificial intelligence (AI). The following section delves into the technical advances in the domain of video analysis and generation.

\subsubsection{Early Approaches}

Generative adversarial networks (GANs) were first applied to generate simple synthetic videos. Models like TiVGAN~\cite{kim2020tivgan} and MoCoGAN~\cite{Tulyakov2018MoCoGAN} pioneered GAN-based video generation.
However, these early GAN models were limited to generating short, low-resolution videos focused on specific domains like human actions. The quality and diversity were lacking.
% Autoregressive Models

\subsubsection{Autoregressive Models}
Compared to GANs, autoregressive models can model density explicitly and conduct stable training, thus they are widely used in visual synthesis.
Autoregressive models~\cite{weissenborn2019scaling, le2021ccvs, wu2022nuwa, ge2022long} tried to generate higher-resolution videos by modeling pixel distributions sequentially. But they were slow and hard to scale up.

\subsubsection{Diffusion Models}
As with still images, Diffusion Models have become very popular for high-quality image generation. Video Diffusion Models (VDM)~\cite{ho2022video} extended image diffusion models to the video domain by training on both images and videos.
Imagen Video~\cite{ho2022imagen} built a cascade of VDMs to generate longer, high-resolution videos. However, it requires large-scale training and latent optimization.
Tune-A-Video~\cite{wu2023tune} optimized the latent space of a diffusion model on a single reference video to adapt it for video generation. This reduced training but still requires optimization.
A recent study by Text2Video-Zero~\cite{khachatryan2023text2video} proposes a zero-shot text-to-video approach without any training on video data. It leverages a pre-trained text-to-image diffusion model and modifies it with motion dynamics in latent space for background consistency and cross-frame attention to preserve foreground details. This allows high-quality video generation from text without costly training. It also enables applications like controlled/specialized video generation and text-driven video editing. Ablations show the contributions of the modifications for temporal consistency. The zero-shot ability and lack of training are advantages over prior techniques.

\begin{figure}
  \begin{center}    \includegraphics[width=0.9\textwidth]{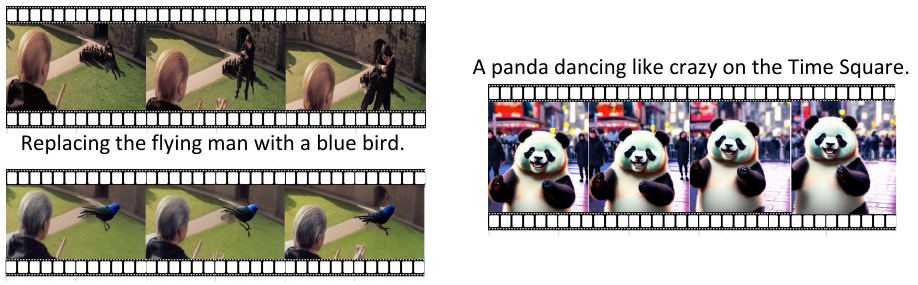}
  \end{center}
  \vspace{0cm}
  \caption{Some examples of Video Generation. \textbf{Left}: Editing a movie with the prompt. 
  \textbf{Right}: Video created by DALL-E 3 with the prompt.
  }\label{fig:overall}
  \vspace{0cm}
\end{figure} 

\subsection{Real-world Applications}
\subsubsection{Film Industry}
New AI technologies such as LLMs or Multi-Modal FMs have the potential to revolutionize the film industry at different stages of the movie-making process~\cite{gpt4film, huang2020movienet}. First, LLMs can analyze the draft scripts and generate unique storylines~\cite{Xiong_2019_ICCV}, which help filmmakers write and revise scripts more efficiently ~\cite{GAIHolly}. 
In addition to scriptwriting, LLMs also can simplify the movie pre-production process~\cite{Huang_2018_CVPR, huang2018person, xia2020online}. Specifically, they can make shooting schedules, find exterior film locations and props, speed up casting person search, and estimate the success and potential revenues the film may earn.
Second, LLMs can generate instructions for technical staff during filming~\cite{huang2018trailers, rao2020unified}, including lighting, shot prediction, audio recording, etc. LLMs are capable of identifying the director's personalized filming style and thus can generate filming instructions specific to the director's style.
Third, Multi-modal LLMs serve as a good editing tool in post-production. LLMs can synthesize multiple clips and even create special effects based on the scripts~\cite{rao2020local}. They can also generate trailers and synopses for promotion purposes~\cite{huang2018trailers, Xiong_2019_ICCV}. LLM-based music composition tools can also be used to find or create an Original Sound Track (OST) that adapts to the movie plot. 

\subsubsection{Social Media}
Increasingly, social media is shifting towards video content over text-based posts. 
From Podcasts to short-form videos (e.g., TikTok) to longer-form user-generated content (e.g. YouTube), users both create and consume video content at ever-growing rates. AI and AIGC are in the early stages of disrupting this industry, but in the near future, this disruption is likely to grow rapidly. One of the most interesting nascent applications is in high-quality AI-based language translation. Several startups have popped up recently that ingest video in a given language (e.g., English) and reproduce it in any number of output languages (e.g., Spanish or Mandarin). The output video can match the original creator's voice characteristics and even make the lips move as if the creator natively speaks the output language\footnote{}. 

\subsubsection{Journalism and Communications}
As illustrated in Sec 2.2, LLMs can analyze large amounts of textual data, including news, social media, and advertisements. Researchers can use LLMs to study how information spreads and impacts the public~\cite{aionmedia}. Video is also an important modality in communication. Nowadays, short user-generated videos are gaining popularity, alongside traditional media like TV and newspapers~\cite{ailocalnews}. The multi-modal FMs have the advantage of analyzing vast amounts of news data in different modalities, including a large quantity of information uploaded by the public. This helps researchers understand how information spreads in the network and track the personal behavior of each user.

\subsubsection{Music Analysis}
Multi-modal LLMs have been proposed to empower frozen LLMs with the capability of understanding both visual and auditory content in videos~\cite{zhang2023video}. Multi-modal FMs can perceive the gestures and movements of the music performers in a video~\cite{li2017video, li2018skeleton, li2017see, li2019online, heydari2023singnet, sargent2014segmentation}, for example, fingering analysis on piano. Based on the visual perception, the FMs can further understand the content, emotion, and intention of the performance~\cite{killian2001effect, li2018creating, thompson2008audio} and reveal the cultural characteristics. The visual understanding provided by FMs helps musicians improve their performance and composition skills~\cite{duan2019performances}.
In addition to audio analysis, these models can help with the generation of music. Diffusion Models (DMs) have been repurposed from images to audio recently, allowing for original musical creations based on text input. In a similar fashion to how a user can interact with an image-based DM to request a given image, a user can also type in a textual description of a requested audio composition and get a sound file based on this description.

\section{Responsible AGI}

% AI vs humans
\textbf{Is AI Threatening Humanity?}
The popularity of AI-generated content, spanning writing, photography, art, and music, has surged dramatically. However, this meteoric rise has also sparked significant backlash, with some people rejecting AI-generated art and even asserting that its widespread adoption signals potential concerns for humanity.
The question of whether AI is threatening humanity is a complex and debated topic. For example, AI itself is a tool created and controlled by humans, which could automate tedious tasks for humans but could also cause job displacement to human society; AI could generate art works efficiently, but they could not serve as a deeper communicative medium of human experience~\cite{bellaiche2023humans}; AI could improve healthcare but could also pose threats to public safety.
Some essential components of responsible AGI are discussed as below. 
% {\color{red} (Need examples.)}

% an article~\footnote{\url{https://news.ubc.ca/2023/08/21/people-dislike-ai-art-because-it-threatens-their-humanity/}}

\subsection{Factuality} 
Large language models are susceptible to hallucinations~\cite{ji2023survey}, wherein they may produce content that includes non-factual information or deviates from established world knowledge~\cite{borji2023categorical}. This poses challenges in numerous applications, such as legal research and historical studies where factual accuracy is crucial. In addition to natural language processing, factuality-related concerns also extend to the field of computer vision. A typical challenge arises in the form of generative models, such as stable diffusion, struggling to accurately generate realistic human hands with the correct number of fingers~\cite{handsfigers} as well as remote sensing images with correct geographic layout \cite{mai2023opportunities}. 
Non-realistic AI-generated images or videos may pose challenges in engaging viewers emotionally or intellectually compared to traditional ones. 

% @Tianze, I need your help for this paragraph (12 lines).
% Follow this paper and use it to find more papers:
% "A Comprehensive Survey of AI-Generated Content (AIGC): A History of Generative AI from GAN to ChatGPT". Sec 7 TRUSTWORTHY & RESPONSIBLE AIGC
Common strategies to tackle the above issues include factuality evaluation and generation regularization. 
For factuality evaluation in generated content, several typical methods stand out. ROUGE~\cite{lin2004rouge} offers a metric that evaluates the quality of computer-generated summaries by measuring their overlap with human-created reference summaries in terms of n-grams, word sequences, and word pairs. Similarly, BLEU~\cite{papineni2002bleu} provides an automatic machine translation evaluation technique renowned for its high correlation with human evaluations, positioning it as a swift and efficient alternative to more labor-intensive human assessments. In a more recent development, a model-based metric~\cite{goodrich2019assessing} has been introduced, specifically designed to assess the factual accuracy of the generated text, further enhancing and complementing the capabilities of traditional methods like ROUGE~\cite{lin2004rouge} and BLEU~\cite{papineni2002bleu}. For generation regularization, "Truthful AI"~\cite{evans2021truthful} is proposed to focus on enhancing the integrity and accuracy of AI-generated outputs. By setting rigorous standards, the initiative seeks to prevent "negligent falsehoods",  achieved through selected datasets and close human-AI interaction, aligning with societal norms and legal constraints.

\subsection{Public Safety}
Despite the rapid advancement of generative AI technology, such as ChatGPT and Midjourney, which can generate human-like texts, images, and videos, it also raises critical concerns related to public safety, encompassing issues of privacy, cybersecurity, national security, individual harassment, and the potential for machine misuse~\cite{guo2023aigc,rao2023building}.

\begin{itemize}
    \item \textbf{Misinformation.} AIGC such as texts, images, and videos can be used to create and spread false or misleading information, leading to public confusion, panic, or harm~\cite{zhou2023synthetic, agrawal2021survey}. For example, Midjourney can accept prompts like ``a hyper-realistic photograph of a man putting election ballots into a box in Phoenix, Arizona'', and produce high-quality images that could be used to support the news~\cite{fakeNewsImage}. The issue is particularly concerning in areas like public health, elections, and emergencies. 
    In addition, AI-generated deep fake images~\cite{trump,biden} and videos~\cite{deepfakeVideo} can impersonate individuals, including public figures, and spread false or defamatory content. Meanwhile, they can be used to invade individuals' privacy by creating content without their consent, leading to serious ethical and legal implications. 
    Repeated exposure to deceptive AI-generated content can damage reputations, incite social unrest, and erode public trust in authorities. Moreover, the National Geospatial-Intelligence Agency (NGA) also alarmed us with the risk of deep fake satellite images from generative AI being used as a terrifying AI-powered weapon \cite{tucker2019newest,zhao2021deepfakegeography}.

    % @Tianze: I need your help to 1) organize the commented content below to complete this paragraph. 2) find some research papers to support each of the cases.
    \item \textbf{Phishing.} Phishing is a type of cyber-attack where attackers attempt to deceive individuals into revealing sensitive or personal information such as login credentials, credit card numbers, or personal information. AI can be used in various ways to enhance phishing campaigns. 
    
    (i) \textbf{Spear Phishing} is a targeted cyber-attack approach that uses \textit{personalized} details to trick individuals into revealing confidential information~\cite{caputo2013going}. Modern LLMs have the ability to produce convincing human-like texts, which can be used to create personalized spear phishing messages on a large scale and at a low cost. For instance, using advanced models like Anthropic's Claude, a hacker can easily generate 1,000 spear phishing emails for just $\$10$ in less than two hours~\cite{hazell2023large}.

    (ii) \textbf{AI voice cloning} is another noteworthy technology, as nowadays only a short voice sample is needed to create a realistic imitation. For instance, Google's AI system can mimic someone's voice with just a five-second sample~\cite{manyam2022artificial}. This technology can be misused in cases where fake audio is used to impersonate authoritative figures in media settings.

    (iii) \textbf{AI-created phishing websites} benefit from the capabilities of multimodal foundation models. These AI-generated websites not only display a remarkable proficiency in emulating the appearance and functionality of established brands, but they also possess the ability to integrate advanced methodologies that can bypass conventional anti-phishing protocols~\cite{roy2023generating}.
    % Specifically, malicious prompts within ChatGPT can be manipulated to craft high-quality phishing websites. 
    
    \item \textbf{Bias and Discrimination.} AI-generated content can perpetuate biases which can harm marginalized groups and exacerbate social inequalities. A recent study~\cite{fang2023bias} shows that AIGC produced by LLMs are more likely to exhibit notable discrimination against underrepresented population groups, compared to authenticated news articles collected from The New York Times and Reuters, where LLMs are asked to generate new articles with the same headlines as the real news. Another example studies AI in generating marketing content such as email composition, recommender systems, and landing page design, which shows that LLMs could amplify biases (e.g., using ``man hours'' to estimate effort, or using ``chairman'' for gender-neutral roles) in the generated content~\cite{biasMarketingContent}
    % \item \textbf{Harassment.}
    Moreover, recent studies also show that although pre-trained LLMs can be used to solve various geospatial tasks \cite{mai2023opportunities,manvi2023geollm}, they also exhibit geographical and geopolitical bias -- so-called geopolitical favouritism which is defined as the over-amplification of certain country representation (eg. countries with higher GDP, geopolitical stability, military strength, etc) in the generated content \cite{faisal2022geographicbias}.
\end{itemize}

Mitigating safety issues caused by AIGC is still an ongoing challenge that requires a collaborative effort from AI developers, regulators, educators, and the broader society.
% @Tianze, I need your help to provide more detailed examples of how to detect: 1) AIGC, 2) AIGC with misinformation such as fake news, 3) AIGC with biases.
Inspired by ``magic must defeat magic'', given the large volume of web content, researchers have been actively working on developing AI-based classifiers to detect online content produced by AI models~\cite{tang2023science}. As highlighted by the work of Ippolito et al.~\cite{ippolito2019automatic}, they rely on the supervised learning approach. Their study specifically fine-tuned the BERT model~\cite{devlin2018bert} using a mix of texts from human authors and those generated by LLMs. This method magnifies the subtle differences between human and AI-produced writings, thus enhancing the model's capability to pinpoint AI-generated content. 
In the field of misinformation detection, AI also plays a crucial role. Zhou et al.~\cite{zhou2023synthetic} investigated the distinct features of AI-generated misinformation and introduced a theory-guided technique to accumulate such content. This facilitates a systematic comparison between human-authored misinformation and its AI-generated counterpart, aiding in the identification of their inherent differences. On another front, AI models are equipped to detect biases within AIGC. Fang et al.~\cite{fang2023bias} selected articles from reputable, impartial news outlets, such as The New York Times and Reuters. By using headlines from these sources as prompts, they assessed the racial and gender biases in LLM-generated content, comparing it with the original articles to highlight discrepancies.

Another line of research focuses on enhancing AI models to reduce the likelihood of misbehavior. For instance, a recent study found that AIGC produced by ChatGPT exhibits a lower level of bias, in part due to its reinforcement learning from human feedback (RLHF) feature~\cite{fang2023bias}.

%% Spear Phishing: AI can help attackers personalize phishing emails by analyzing social media data and other online information to craft convincing messages that appear to come from trusted sources.

% Automated Messaging: AI chatbots or natural language processing (NLP) models can automate phishing interactions, making it easier for attackers to engage with multiple victims simultaneously.

%% Voice Cloning: AI-generated voice technology can be used to mimic trusted individuals' voices in phishing phone calls, making them sound more convincing.

%% Phishing Website: AI can be used to create realistic-looking phishing websites that imitate legitimate ones. It can also be used for detecting and bypassing security measures.

% Social Engineering: AI can assist in creating more persuasive social engineering narratives, tailoring messages to exploit human emotions, curiosity, or fear.

\subsection{Toxicity}
To ensure the dependable deployment of AI, it is imperative to prevent AI models from generating toxic or harmful content, which encompasses hate speech, biases, cyberbullying, and other objectionable material. Toxic content can harm individuals and communities, perpetuate discrimination, and create a hostile online environment. 
Although detecting hate speech and offensive language has long been a subject of research~\cite{davidson2017automated,zampieri2019semeval}, the
study of toxic AI-generated content is a more recent direction. 
For example, recent findings indicate that ChatGPT can consistently generate toxic content on a broad spectrum of topics when it is assigned a persona~\cite{deshpande2023toxicity}. Pre-trained language models can produce toxic text even when prompted with seemingly innocuous inputs~\cite{gehman2020realtoxicityprompts}.
Thus, many organizations were actively working on research and technology to improve AI content generation while reducing harmful outputs. These recent efforts can be divided into two categories, including training-time and inference-time detoxification.

\textbf{Training-time Strategies.} There are two primary methods for refining large foundation models: \textit{pre-training} and \textit{fine-tuning}. To improve model pre-training, one approach involves the identification and filtering of undesirable documents from the training data~\cite{ngo2021mitigating}. Additionally, we could augment the training data with information pertaining to its toxicity, towards guiding the LM to detect toxic content and hence generate non-toxic text~\cite{prabhumoye2023adding}.
During fine-tuning, it is possible to align language models with human preferences by employing human feedback as a reward signal~\cite{wang2022exploring,ouyang2022training,gururangan2020don}. A well-known example is InstructGPT~\cite{ouyang2022training} developed by OpenAI, which could generate less toxic outputs than those from GPT-3 by using properly designed prompts.

\textbf{Inference-time Strategies.} There are two major methods for reducing the toxicity of AI-generated content during inference time, including prompt learning and decoding-time steering. Prompt learning offers a versatile method to assess and tailor the output of large language models, such as toxicity classification, toxic text span detection, and detoxification~\cite{he2023you}. First, given a sentence, an initial step involves mapping its label to either "Yes" or "No" and subsequently refining the prompt to enhance its guidance for the language model. Second, toxic text span detection identifies the specific segments (i.e., the word offsets) that make the text toxic. Third, to rephrase the toxic text into a non-toxic version while preserving its semantic meaning.
On the other hand, decoding-time steering~\cite{dathathri2019plug, liu2021dexperts, gehman2020realtoxicityprompts} manipulates the output distribution to avoid generating mindless and offensive content.

% \subsection{Privacy}
% Education

\section{Conclusion}
The swift evolution of artificial general intelligence (AGI) is transforming the landscape of art and humanities in profound ways. As demonstrated in this paper, AGI systems like large language models and creative image generators have already exhibited impressive capabilities across diverse artistic domains including literature, visual arts, music, and more. However, as boundaries between human creativity and machine capabilities blur, difficult questions emerge around truth, toxicity, biases, accountability, and social impacts.

While celebrating the immense potential of AGI to augment human expression, we must thoughtfully navigate its responsible development. Multi-stakeholder collaboration and public discourse are vital to steer these systems in directions that uphold cultural values, pluralism, dignity, and truth. Technical solutions such as robust factuality evaluations, toxicity filters, and bias detectors can help instill reliability and trustworthiness in AGI systems. Ultimately, however, cultural shifts toward responsible innovation, centered on human flourishing over profits or progress for its own sake, are crucial.

By harnessing AGI as a partner for human creativity, while proactively addressing its pitfalls, we can usher in an era where machine intelligence promotes knowledge, empowers imagination, and expands access to the arts. The onus lies on researchers, developers, policymakers, and society at large to align AGI's technological promise with enduring human values. Through principled efforts, we can ensure these rapidly evolving systems enrich rather than undermine our shared cultural heritage.

\paragraph{Acknowledgement}
We would like to thank Prof. John Hale from the Linguistics Department, University of Georgia for his thoughtful comments on the opportunities of AIGC and foundation models' applications on various art and humanities tasks.

\bibliographystyle{unsrt}

\begin{thebibliography}{100}
	
	\bibitem{pool2018looking}
	Natalie~M Pool.
	\newblock Looking inward: Philosophical and methodological perspectives on
	phenomenological self-reflection.
	\newblock {\em Nursing science quarterly}, 31(3):245--252, 2018.
	
	\bibitem{ambartsoumean2023ai}
	Vemir~Michael Ambartsoumean and Roman~V Yampolskiy.
	\newblock Ai risk skepticism, a comprehensive survey.
	\newblock {\em arXiv preprint arXiv:2303.03885}, 2023.
	
	\bibitem{ramesh2021zero}
	Aditya Ramesh, Mikhail Pavlov, Gabriel Goh, Scott Gray, Chelsea Voss, Alec
	Radford, Mark Chen, and Ilya Sutskever.
	\newblock Zero-shot text-to-image generation.
	\newblock In {\em International Conference on Machine Learning}, pages
	8821--8831. PMLR, 2021.
	
	\bibitem{mordvintsev2015deepdream}
	Alexander Mordvintsev, Christopher Olah, and Mike Tyka.
	\newblock Deepdream-a code example for visualizing neural networks.
	\newblock {\em Google Research}, 2(5), 2015.
	
	\bibitem{singh2021neural}
	Akhil Singh, Vaibhav Jaiswal, Gaurav Joshi, Adith Sanjeeve, Shilpa Gite, and
	Ketan Kotecha.
	\newblock Neural style transfer: A critical review.
	\newblock {\em IEEE Access}, 9:131583--131613, 2021.
	
	\bibitem{goodfellow2020gan}
	Ian Goodfellow, Jean Pouget-Abadie, Mehdi Mirza, Bing Xu, David Warde-Farley,
	Sherjil Ozair, Aaron Courville, and Yoshua Bengio.
	\newblock Generative adversarial networks.
	\newblock {\em Communications of the ACM}, 63(11):139--144, 2020.
	
	\bibitem{cohn2018nyt}
	Gabe Cohn.
	\newblock Ai art at christie’s sells for \$432,500.
	\newblock {\em New York Times}, 2018.
	
	\bibitem{esser2021taming}
	Patrick Esser, Robin Rombach, and Bjorn Ommer.
	\newblock Taming transformers for high-resolution image synthesis.
	\newblock In {\em Proceedings of the IEEE/CVF conference on computer vision and
		pattern recognition}, pages 12873--12883, 2021.
	
	\bibitem{radford2021learning}
	Alec Radford, Jong~Wook Kim, Chris Hallacy, Aditya Ramesh, Gabriel Goh,
	Sandhini Agarwal, Girish Sastry, Amanda Askell, Pamela Mishkin, Jack Clark,
	et~al.
	\newblock Learning transferable visual models from natural language
	supervision.
	\newblock In {\em International conference on machine learning}, pages
	8748--8763. PMLR, 2021.
	
	\bibitem{rombach2022high}
	Robin Rombach, Andreas Blattmann, Dominik Lorenz, Patrick Esser, and Bj{\"o}rn
	Ommer.
	\newblock High-resolution image synthesis with latent diffusion models.
	\newblock In {\em Proceedings of the IEEE/CVF conference on computer vision and
		pattern recognition}, pages 10684--10695, 2022.
	
	\bibitem{saharia2022imagen}
	Chitwan Saharia, William Chan, Saurabh Saxena, Lala Li, Jay Whang, Emily~L
	Denton, Kamyar Ghasemipour, Raphael Gontijo~Lopes, Burcu Karagol~Ayan, Tim
	Salimans, et~al.
	\newblock Photorealistic text-to-image diffusion models with deep language
	understanding.
	\newblock {\em Advances in Neural Information Processing Systems},
	35:36479--36494, 2022.
	
	\bibitem{zhao2023brain}
	Lin Zhao, Lu~Zhang, Zihao Wu, Yuzhong Chen, Haixing Dai, Xiaowei Yu, Zhengliang
	Liu, Tuo Zhang, Xintao Hu, Xi~Jiang, et~al.
	\newblock When brain-inspired ai meets agi.
	\newblock {\em Meta-Radiology}, page 100005, 2023.
	
	\bibitem{cao2023comprehensive}
	Yihan Cao, Siyu Li, Yixin Liu, Zhiling Yan, Yutong Dai, Philip~S Yu, and Lichao
	Sun.
	\newblock A comprehensive survey of ai-generated content (aigc): A history of
	generative ai from gan to chatgpt.
	\newblock {\em arXiv preprint arXiv:2303.04226}, 2023.
	
	\bibitem{brown2020language}
	Tom Brown, Benjamin Mann, Nick Ryder, Melanie Subbiah, Jared~D Kaplan, Prafulla
	Dhariwal, Arvind Neelakantan, Pranav Shyam, Girish Sastry, Amanda Askell,
	et~al.
	\newblock Language models are few-shot learners.
	\newblock {\em Advances in neural information processing systems},
	33:1877--1901, 2020.
	
	\bibitem{liu2023summary}
	Yiheng Liu, Tianle Han, Siyuan Ma, Jiayue Zhang, Yuanyuan Yang, Jiaming Tian,
	Hao He, Antong Li, Mengshen He, Zhengliang Liu, et~al.
	\newblock Summary of chatgpt-related research and perspective towards the
	future of large language models.
	\newblock {\em Meta-Radiology}, page 100017, 2023.
	
	\bibitem{openai2023gpt}
	OpenAI.
	\newblock Gpt-4 technical report.
	\newblock {\em arXiv}, pages 2303--08774, 2023.
	
	\bibitem{lewis2019bart}
	Mike Lewis, Yinhan Liu, Naman Goyal, Marjan Ghazvininejad, Abdelrahman Mohamed,
	Omer Levy, Ves Stoyanov, and Luke Zettlemoyer.
	\newblock Bart: Denoising sequence-to-sequence pre-training for natural
	language generation, translation, and comprehension.
	\newblock {\em arXiv preprint arXiv:1910.13461}, 2019.
	
	\bibitem{raffel2020exploring}
	Colin Raffel, Noam Shazeer, Adam Roberts, Katherine Lee, Sharan Narang, Michael
	Matena, Yanqi Zhou, Wei Li, and Peter~J Liu.
	\newblock Exploring the limits of transfer learning with a unified text-to-text
	transformer.
	\newblock {\em The Journal of Machine Learning Research}, 21(1):5485--5551,
	2020.
	
	\bibitem{ramesh2022hierarchical}
	Aditya Ramesh, Prafulla Dhariwal, Alex Nichol, Casey Chu, and Mark Chen.
	\newblock Hierarchical text-conditional image generation with clip latents.
	\newblock {\em arXiv preprint arXiv:2204.06125}, 1(2):3, 2022.
	
	\bibitem{nichol2021glide}
	Alex Nichol, Prafulla Dhariwal, Aditya Ramesh, Pranav Shyam, Pamela Mishkin,
	Bob McGrew, Ilya Sutskever, and Mark Chen.
	\newblock Glide: Towards photorealistic image generation and editing with
	text-guided diffusion models.
	\newblock {\em arXiv preprint arXiv:2112.10741}, 2021.
	
	\bibitem{wu2023ai}
	Jiayang Wu, Wensheng Gan, Zefeng Chen, Shicheng Wan, and Hong Lin.
	\newblock Ai-generated content (aigc): A survey.
	\newblock {\em arXiv preprint arXiv:2304.06632}, 2023.
	
	\bibitem{chen2023large}
	Jin Chen, Zheng Liu, Xu~Huang, Chenwang Wu, Qi~Liu, Gangwei Jiang, Yuanhao Pu,
	Yuxuan Lei, Xiaolong Chen, Xingmei Wang, et~al.
	\newblock When large language models meet personalization: Perspectives of
	challenges and opportunities.
	\newblock {\em arXiv preprint arXiv:2307.16376}, 2023.
	
	\bibitem{liu3surviving}
	Zhengliang Liu, Lu~Zhang, Zihao Wu, Xiaowei Yu, Chao Cao, Haixing Dai, Ninghao
	Liu, Jun Liu, Wei Liu, Quanzheng Li, et~al.
	\newblock Surviving chatgpt in healthcare.
	\newblock {\em Frontiers in Radiology}, 3:1224682.
	
	\bibitem{vaswani2017attention}
	Ashish Vaswani, Noam Shazeer, Niki Parmar, Jakob Uszkoreit, Llion Jones,
	Aidan~N Gomez, {\L}ukasz Kaiser, and Illia Polosukhin.
	\newblock Attention is all you need.
	\newblock {\em Advances in neural information processing systems}, 30, 2017.
	
	\bibitem{sherstinsky2020fundamentals}
	Alex Sherstinsky.
	\newblock Fundamentals of recurrent neural network (rnn) and long short-term
	memory (lstm) network.
	\newblock {\em Physica D: Nonlinear Phenomena}, 404:132306, 2020.
	
	\bibitem{staudemeyer2019understanding}
	Ralf~C Staudemeyer and Eric~Rothstein Morris.
	\newblock Understanding lstm--a tutorial into long short-term memory recurrent
	neural networks.
	\newblock {\em arXiv preprint arXiv:1909.09586}, 2019.
	
	\bibitem{gehring2017convolutional}
	Jonas Gehring, Michael Auli, David Grangier, Denis Yarats, and Yann~N Dauphin.
	\newblock Convolutional sequence to sequence learning.
	\newblock In {\em International conference on machine learning}, pages
	1243--1252. PMLR, 2017.
	
	\bibitem{hou2007saliency}
	Xiaodi Hou and Liqing Zhang.
	\newblock Saliency detection: A spectral residual approach.
	\newblock In {\em 2007 IEEE Conference on computer vision and pattern
		recognition}, pages 1--8. Ieee, 2007.
	
	\bibitem{devlin2018bert}
	Jacob Devlin, Ming-Wei Chang, Kenton Lee, and Kristina Toutanova.
	\newblock Bert: Pre-training of deep bidirectional transformers for language
	understanding.
	\newblock {\em arXiv preprint arXiv:1810.04805}, 2018.
	
	\bibitem{meyerson2017beyond}
	Elliot Meyerson and Risto Miikkulainen.
	\newblock Beyond shared hierarchies: Deep multitask learning through soft layer
	ordering.
	\newblock {\em arXiv preprint arXiv:1711.00108}, 2017.
	
	\bibitem{freitag2017beam}
	Markus Freitag and Yaser Al-Onaizan.
	\newblock Beam search strategies for neural machine translation.
	\newblock {\em arXiv preprint arXiv:1702.01806}, 2017.
	
	\bibitem{gu2017trainable}
	Jiatao Gu, Kyunghyun Cho, and Victor~OK Li.
	\newblock Trainable greedy decoding for neural machine translation.
	\newblock {\em arXiv preprint arXiv:1702.02429}, 2017.
	
	\bibitem{yu2022scaling}
	Jiahui Yu, Yuanzhong Xu, Jing~Yu Koh, Thang Luong, Gunjan Baid, Zirui Wang,
	Vijay Vasudevan, Alexander Ku, Yinfei Yang, Burcu~Karagol Ayan, et~al.
	\newblock Scaling autoregressive models for content-rich text-to-image
	generation.
	\newblock {\em arXiv preprint arXiv:2206.10789}, 2(3):5, 2022.
	
	\bibitem{radford2018improving}
	Alec Radford, Karthik Narasimhan, Tim Salimans, Ilya Sutskever, et~al.
	\newblock Improving language understanding by generative pre-training.
	\newblock 2018.
	
	\bibitem{kaddour2023challenges}
	Jean Kaddour, Joshua Harris, Maximilian Mozes, Herbie Bradley, Roberta
	Raileanu, and Robert McHardy.
	\newblock Challenges and applications of large language models.
	\newblock {\em arXiv preprint arXiv:2307.10169}, 2023.
	
	\bibitem{li2022survey}
	Huayang Li, Yixuan Su, Deng Cai, Yan Wang, and Lemao Liu.
	\newblock A survey on retrieval-augmented text generation.
	\newblock {\em arXiv preprint arXiv:2202.01110}, 2022.
	
	\bibitem{zhao2023survey}
	Wayne~Xin Zhao, Kun Zhou, Junyi Li, Tianyi Tang, Xiaolei Wang, Yupeng Hou,
	Yingqian Min, Beichen Zhang, Junjie Zhang, Zican Dong, et~al.
	\newblock A survey of large language models.
	\newblock {\em arXiv preprint arXiv:2303.18223}, 2023.
	
	\bibitem{kasneci2023chatgpt}
	Enkelejda Kasneci, Kathrin Se{\ss}ler, Stefan K{\"u}chemann, Maria Bannert,
	Daryna Dementieva, Frank Fischer, Urs Gasser, Georg Groh, Stephan
	G{\"u}nnemann, Eyke H{\"u}llermeier, et~al.
	\newblock Chatgpt for good? on opportunities and challenges of large language
	models for education.
	\newblock {\em Learning and individual differences}, 103:102274, 2023.
	
	\bibitem{chang2023survey}
	Yupeng Chang, Xu~Wang, Jindong Wang, Yuan Wu, Kaijie Zhu, Hao Chen, Linyi Yang,
	Xiaoyuan Yi, Cunxiang Wang, Yidong Wang, et~al.
	\newblock A survey on evaluation of large language models.
	\newblock {\em arXiv preprint arXiv:2307.03109}, 2023.
	
	\bibitem{wei2022emergent}
	Jason Wei, Yi~Tay, Rishi Bommasani, Colin Raffel, Barret Zoph, Sebastian
	Borgeaud, Dani Yogatama, Maarten Bosma, Denny Zhou, Donald Metzler, et~al.
	\newblock Emergent abilities of large language models.
	\newblock {\em arXiv preprint arXiv:2206.07682}, 2022.
	
	\bibitem{chen2021evaluating}
	Mark Chen, Jerry Tworek, Heewoo Jun, Qiming Yuan, Henrique Ponde de~Oliveira
	Pinto, Jared Kaplan, Harri Edwards, Yuri Burda, Nicholas Joseph, Greg
	Brockman, et~al.
	\newblock Evaluating large language models trained on code.
	\newblock {\em arXiv preprint arXiv:2107.03374}, 2021.
	
	\bibitem{linzen2019can}
	Tal Linzen.
	\newblock What can linguistics and deep learning contribute to each other?
	response to pater.
	\newblock {\em Language}, 95(1):e99--e108, 2019.
	
	\bibitem{baroni2022proper}
	Marco Baroni.
	\newblock On the proper role of linguistically-oriented deep net analysis in
	linguistic theorizing.
	\newblock {\em Algebraic structures in natural language}, pages 1--16, 2022.
	
	\bibitem{linzen2021syntactic}
	Tal Linzen and Marco Baroni.
	\newblock Syntactic structure from deep learning.
	\newblock {\em Annual Review of Linguistics}, 7:195--212, 2021.
	
	\bibitem{katzir2023large}
	Roni Katzir.
	\newblock Why large language models are poor theories of human linguistic
	cognition. a reply to piantadosi (2023).
	\newblock {\em Manuscript. Tel Aviv University. url: https://lingbuzz.
		net/lingbuzz/007190}, 2023.
	
	\bibitem{zhang2023sentiment}
	Wenxuan Zhang, Yue Deng, Bing Liu, Sinno~Jialin Pan, and Lidong Bing.
	\newblock Sentiment analysis in the era of large language models: A reality
	check.
	\newblock {\em arXiv preprint arXiv:2305.15005}, 2023.
	
	\bibitem{dai2023ad}
	Haixing Dai, Yiwei Li, Zhengliang Liu, Lin Zhao, Zihao Wu, Suhang Song,
	Ye~Shen, Dajiang Zhu, Xiang Li, Sheng Li, et~al.
	\newblock Ad-autogpt: An autonomous gpt for alzheimer's disease infodemiology.
	\newblock {\em arXiv preprint arXiv:2306.10095}, 2023.
	
	\bibitem{he2012generating}
	Jing He, Ming Zhou, and Long Jiang.
	\newblock Generating chinese classical poems with statistical machine
	translation models.
	\newblock In {\em Proceedings of the AAAI Conference on Artificial
		Intelligence}, volume~26, pages 1650--1656, 2012.
	
	\bibitem{zelch2023commercialized}
	Ines Zelch, Matthias Hagen, and Martin Potthast.
	\newblock Commercialized generative ai: A critical study of the feasibility and
	ethics of generating native advertising using large language models in
	conversational web search.
	\newblock {\em arXiv preprint arXiv:2310.04892}, 2023.
	
	\bibitem{rivas2023marketing}
	Pablo Rivas and Liang Zhao.
	\newblock Marketing with chatgpt: Navigating the ethical terrain of gpt-based
	chatbot technology.
	\newblock {\em AI}, 4(2):375--384, 2023.
	
	\bibitem{goodfellow2014generative}
	Ian Goodfellow, Jean Pouget-Abadie, Mehdi Mirza, Bing Xu, David Warde-Farley,
	Sherjil Ozair, Aaron Courville, and Yoshua Bengio.
	\newblock Generative adversarial nets.
	\newblock {\em Advances in neural information processing systems}, 27, 2014.
	
	\bibitem{mirza2014cgan}
	Mehdi Mirza and Simon Osindero.
	\newblock Conditional generative adversarial nets.
	\newblock {\em arXiv preprint arXiv:1411.1784}, 2014.
	
	\bibitem{pu2016vae}
	Yunchen Pu, Zhe Gan, Ricardo Henao, Xin Yuan, Chunyuan Li, Andrew Stevens, and
	Lawrence Carin.
	\newblock Variational autoencoder for deep learning of images, labels and
	captions.
	\newblock {\em Advances in neural information processing systems}, 29, 2016.
	
	\bibitem{kobyzev2020normalizing}
	Ivan Kobyzev, Simon~JD Prince, and Marcus~A Brubaker.
	\newblock Normalizing flows: An introduction and review of current methods.
	\newblock {\em IEEE transactions on pattern analysis and machine intelligence},
	43(11):3964--3979, 2020.
	
	\bibitem{li2018point}
	Chun-Liang Li, Manzil Zaheer, Yang Zhang, Barnabas Poczos, and Ruslan
	Salakhutdinov.
	\newblock Point cloud gan.
	\newblock {\em arXiv preprint arXiv:1810.05795}, 2018.
	
	\bibitem{shu20193d}
	Dong~Wook Shu, Sung~Woo Park, and Junseok Kwon.
	\newblock 3d point cloud generative adversarial network based on tree
	structured graph convolutions.
	\newblock In {\em Proceedings of the IEEE/CVF international conference on
		computer vision}, pages 3859--3868, 2019.
	
	\bibitem{ramasinghe2020spectral}
	Sameera Ramasinghe, Salman Khan, Nick Barnes, and Stephen Gould.
	\newblock Spectral-gans for high-resolution 3d point-cloud generation.
	\newblock In {\em 2020 IEEE/RSJ International Conference on Intelligent Robots
		and Systems (IROS)}, pages 8169--8176. IEEE, 2020.
	
	\bibitem{wang2018graphgan}
	Hongwei Wang, Jia Wang, Jialin Wang, Miao Zhao, Weinan Zhang, Fuzheng Zhang,
	Xing Xie, and Minyi Guo.
	\newblock Graphgan: Graph representation learning with generative adversarial
	nets.
	\newblock In {\em Proceedings of the AAAI conference on artificial
		intelligence}, volume~32, 2018.
	
	\bibitem{wu2016learning}
	Jiajun Wu, Chengkai Zhang, Tianfan Xue, Bill Freeman, and Josh Tenenbaum.
	\newblock Learning a probabilistic latent space of object shapes via 3d
	generative-adversarial modeling.
	\newblock {\em Advances in neural information processing systems}, 29, 2016.
	
	\bibitem{jing2019neural}
	Yongcheng Jing, Yezhou Yang, Zunlei Feng, Jingwen Ye, Yizhou Yu, and Mingli
	Song.
	\newblock Neural style transfer: A review.
	\newblock {\em IEEE transactions on visualization and computer graphics},
	26(11):3365--3385, 2019.
	
	\bibitem{doersch2016tutorial}
	Carl Doersch.
	\newblock Tutorial on variational autoencoders.
	\newblock {\em arXiv preprint arXiv:1606.05908}, 2016.
	
	\bibitem{mansimov2015generating}
	Elman Mansimov, Emilio Parisotto, Jimmy~Lei Ba, and Ruslan Salakhutdinov.
	\newblock Generating images from captions with attention.
	\newblock {\em arXiv preprint arXiv:1511.02793}, 2015.
	
	\bibitem{nobari2021creativegan}
	Amin~Heyrani Nobari, Muhammad~Fathy Rashad, and Faez Ahmed.
	\newblock Creativegan: Editing generative adversarial networks for creative
	design synthesis.
	\newblock {\em arXiv preprint arXiv:2103.06242}, 2021.
	
	\bibitem{vondrick2016generating}
	Carl Vondrick, Hamed Pirsiavash, and Antonio Torralba.
	\newblock Generating videos with scene dynamics.
	\newblock {\em Advances in neural information processing systems}, 29, 2016.
	
	\bibitem{arnab2021vivit}
	Anurag Arnab, Mostafa Dehghani, Georg Heigold, Chen Sun, Mario Lu{\v{c}}i{\'c},
	and Cordelia Schmid.
	\newblock Vivit: A video vision transformer.
	\newblock In {\em Proceedings of the IEEE/CVF international conference on
		computer vision}, pages 6836--6846, 2021.
	
	\bibitem{liu2022video}
	Ze~Liu, Jia Ning, Yue Cao, Yixuan Wei, Zheng Zhang, Stephen Lin, and Han Hu.
	\newblock Video swin transformer.
	\newblock In {\em Proceedings of the IEEE/CVF conference on computer vision and
		pattern recognition}, pages 3202--3211, 2022.
	
	\bibitem{xu2018attngan}
	Tao Xu, Pengchuan Zhang, Qiuyuan Huang, Han Zhang, Zhe Gan, Xiaolei Huang, and
	Xiaodong He.
	\newblock Attngan: Fine-grained text to image generation with attentional
	generative adversarial networks.
	\newblock In {\em Proceedings of the IEEE conference on computer vision and
		pattern recognition}, pages 1316--1324, 2018.
	
	\bibitem{tao2022df}
	Ming Tao, Hao Tang, Fei Wu, Xiao-Yuan Jing, Bing-Kun Bao, and Changsheng Xu.
	\newblock Df-gan: A simple and effective baseline for text-to-image synthesis.
	\newblock In {\em Proceedings of the IEEE/CVF Conference on Computer Vision and
		Pattern Recognition}, pages 16515--16525, 2022.
	
	\bibitem{sohl2015deep}
	Jascha Sohl-Dickstein, Eric Weiss, Niru Maheswaranathan, and Surya Ganguli.
	\newblock Deep unsupervised learning using nonequilibrium thermodynamics.
	\newblock In {\em International conference on machine learning}, pages
	2256--2265. PMLR, 2015.
	
	\bibitem{ho2020denoising}
	Jonathan Ho, Ajay Jain, and Pieter Abbeel.
	\newblock Denoising diffusion probabilistic models.
	\newblock {\em Advances in neural information processing systems},
	33:6840--6851, 2020.
	
	\bibitem{nichol2021improved}
	Alexander~Quinn Nichol and Prafulla Dhariwal.
	\newblock Improved denoising diffusion probabilistic models.
	\newblock In {\em International Conference on Machine Learning}, pages
	8162--8171. PMLR, 2021.
	
	\bibitem{song2020score}
	Yang Song, Jascha Sohl-Dickstein, Diederik~P Kingma, Abhishek Kumar, Stefano
	Ermon, and Ben Poole.
	\newblock Score-based generative modeling through stochastic differential
	equations.
	\newblock {\em arXiv preprint arXiv:2011.13456}, 2020.
	
	\bibitem{song2020denoising}
	Jiaming Song, Chenlin Meng, and Stefano Ermon.
	\newblock Denoising diffusion implicit models.
	\newblock {\em arXiv preprint arXiv:2010.02502}, 2020.
	
	\bibitem{dhariwal2021diffusion}
	Prafulla Dhariwal and Alexander Nichol.
	\newblock Diffusion models beat gans on image synthesis.
	\newblock {\em Advances in neural information processing systems},
	34:8780--8794, 2021.
	
	\bibitem{ho2022classifier}
	Jonathan Ho and Tim Salimans.
	\newblock Classifier-free diffusion guidance.
	\newblock {\em arXiv preprint arXiv:2207.12598}, 2022.
	
	\bibitem{saharia2022image}
	Chitwan Saharia, Jonathan Ho, William Chan, Tim Salimans, David~J Fleet, and
	Mohammad Norouzi.
	\newblock Image super-resolution via iterative refinement.
	\newblock {\em IEEE Transactions on Pattern Analysis and Machine Intelligence},
	45(4):4713--4726, 2022.
	
	\bibitem{kingma2019introduction}
	Diederik~P Kingma, Max Welling, et~al.
	\newblock An introduction to variational autoencoders.
	\newblock {\em Foundations and Trends{\textregistered} in Machine Learning},
	12(4):307--392, 2019.
	
	\bibitem{krygier1995cartography}
	John~B Krygier.
	\newblock Cartography as an art and a science?
	\newblock {\em The Cartographic Journal}, 32(1):3--10, 1995.
	
	\bibitem{mulcahy2001symbolization}
	Karen~A Mulcahy and Keith~C Clarke.
	\newblock Symbolization of map projection distortion: a review.
	\newblock {\em Cartography and geographic information science}, 28(3):167--182,
	2001.
	
	\bibitem{chrisman2017calculating}
	Nicholas~R Chrisman.
	\newblock Calculating on a round planet.
	\newblock {\em International Journal of Geographical Information Science},
	31(4):637--657, 2017.
	
	\bibitem{mai2023sphere2vec}
	Gengchen Mai, Yao Xuan, Wenyun Zuo, Yutong He, Jiaming Song, Stefano Ermon,
	Krzysztof Janowicz, and Ni~Lao.
	\newblock Sphere2vec: A general-purpose location representation learning over a
	spherical surface for large-scale geospatial predictions.
	\newblock {\em ISPRS Journal of Photogrammetry and Remote Sensing},
	202:439--462, 2023.
	
	\bibitem{brassel1988review}
	Kurt~E Brassel and Robert Weibel.
	\newblock A review and conceptual framework of automated map generalization.
	\newblock {\em International Journal of Geographical Information System},
	2(3):229--244, 1988.
	
	\bibitem{ai2001map}
	Tinghua Ai and Peter van Oosterom.
	\newblock A map generalization model based on algebra mapping transformation.
	\newblock In {\em Proceedings of the 9th ACM international symposium on
		Advances in geographic information systems}, pages 21--27, 2001.
	
	\bibitem{kang2020towards}
	Y~Kang, J~Rao, W~Wang, B~Peng, S~Gao, and F~Zhang.
	\newblock Towards cartographic knowledge encoding with deep learning: A case
	study of building generalization.
	\newblock In {\em Proceedings of the AutoCarto}, 2020.
	
	\bibitem{he2018recognition}
	Xianjin He, Xinchang Zhang, and Qinchuan Xin.
	\newblock Recognition of building group patterns in topographic maps based on
	graph partitioning and random forest.
	\newblock {\em ISPRS Journal of Photogrammetry and Remote Sensing}, 136:26--40,
	2018.
	
	\bibitem{yan2019graph}
	Xiongfeng Yan, Tinghua Ai, Min Yang, and Hongmei Yin.
	\newblock A graph convolutional neural network for classification of building
	patterns using spatial vector data.
	\newblock {\em ISPRS journal of photogrammetry and remote sensing},
	150:259--273, 2019.
	
	\bibitem{mai2023towards}
	Gengchen Mai, Chiyu Jiang, Weiwei Sun, Rui Zhu, Yao Xuan, Ling Cai, Krzysztof
	Janowicz, Stefano Ermon, and Ni~Lao.
	\newblock Towards general-purpose representation learning of polygonal
	geometries.
	\newblock {\em GeoInformatica}, 27(2):289--340, 2023.
	
	\bibitem{yu2022recognition}
	Huafei Yu, Tinghua Ai, Min Yang, Lina Huang, and Jiaming Yuan.
	\newblock A recognition method for drainage patterns using a graph
	convolutional network.
	\newblock {\em International Journal of Applied Earth Observation and
		Geoinformation}, 107:102696, 2022.
	
	\bibitem{mai2019space2vec}
	Gengchen Mai, Krzysztof Janowicz, Bo~Yan, Rui Zhu, Ling Cai, and Ni~Lao.
	\newblock Multi-scale representation learning for spatial feature distributions
	using grid cells.
	\newblock In {\em International Conference on Learning Representations}, 2020.
	
	\bibitem{mai2023csp}
	Gengchen Mai, Ni~Lao, Yutong He, Jiaming Song, and Stefano Ermon.
	\newblock Csp: Self-supervised contrastive spatial pre-training for
	geospatial-visual representations.
	\newblock In {\em the Fortieth International Conference on Machine Learning
		(ICML 2023)}, 2023.
	
	\bibitem{mai2022review}
	Gengchen Mai, Krzysztof Janowicz, Yingjie Hu, Song Gao, Bo~Yan, Rui Zhu, Ling
	Cai, and Ni~Lao.
	\newblock A review of location encoding for geoai: methods and applications.
	\newblock {\em International Journal of Geographical Information Science},
	36(4):639--673, 2022.
	
	\bibitem{mai2022towards}
	Gengchen Mai, Chris Cundy, Kristy Choi, Yingjie Hu, Ni~Lao, and Stefano Ermon.
	\newblock Towards a foundation model for geospatial artificial intelligence
	(vision paper).
	\newblock In {\em Proceedings of the 30th International Conference on Advances
		in Geographic Information Systems}, pages 1--4, 2022.
	
	\bibitem{mai2023opportunities}
	Gengchen Mai, Weiming Huang, Jin Sun, Suhang Song, Deepak Mishra, Ninghao Liu,
	Song Gao, Tianming Liu, Gao Cong, Yingjie Hu, et~al.
	\newblock On the opportunities and challenges of foundation models for
	geospatial artificial intelligence.
	\newblock {\em arXiv preprint arXiv:2304.06798}, 2023.
	
	\bibitem{peng2023kosmos2}
	Zhiliang Peng, Wenhui Wang, Li~Dong, Yaru Hao, Shaohan Huang, Shuming Ma, and
	Furu Wei.
	\newblock Kosmos-2: Grounding multimodal large language models to the world.
	\newblock {\em arXiv preprint arXiv:2306.14824}, 2023.
	
	\bibitem{shbita2023building}
	Basel Shbita, Craig~A Knoblock, Weiwei Duan, Yao-Yi Chiang, Johannes~H Uhl, and
	Stefan Leyk.
	\newblock Building spatio-temporal knowledge graphs from vectorized topographic
	historical maps.
	\newblock {\em Semantic Web}, (Preprint):1--23, 2023.
	
	\bibitem{kim2023mapkurator}
	Jina Kim, Zekun Li, Yijun Lin, Min Namgung, Leeje Jang, and Yao-Yi Chiang.
	\newblock The mapkurator system: A complete pipeline for extracting and linking
	text from historical maps.
	\newblock {\em arXiv preprint arXiv:2306.17059}, 2023.
	
	\bibitem{mai2022narrative}
	Gengchen Mai, Weiming Huang, Ling Cai, Rui Zhu, and Ni~Lao.
	\newblock Narrative cartography with knowledge graphs.
	\newblock {\em Journal of Geovisualization and Spatial Analysis}, 6(1):4, 2022.
	
	\bibitem{kang2023ethics}
	Yuhao Kang, Qianheng Zhang, and Robert Roth.
	\newblock The ethics of ai-generated maps: A study of dalle 2 and implications
	for cartography.
	\newblock {\em arXiv preprint arXiv:2304.10743}, 2023.
	
	\bibitem{fernberg2023artificial}
	Phillip Fernberg and Brent Chamberlain.
	\newblock Artificial intelligence in landscape architecture: A literature
	review.
	\newblock {\em Landscape Journal}, 42(1):13--35, 2023.
	
	\bibitem{li2023designer}
	Marika Li.
	\newblock {\em Designer Robots: An early look at applications for Artificial
		Intelligence Visualization Software in Landscape Architecture}.
	\newblock PhD thesis, University of Guelph, 2023.
	
	\bibitem{seneviratne2022dalle}
	Sachith Seneviratne, Damith Senanayake, Sanka Rasnayaka, Rajith Vidanaarachchi,
	and Jason Thompson.
	\newblock Dalle-urban: Capturing the urban design expertise of large text to
	image transformers.
	\newblock In {\em 2022 International Conference on Digital Image Computing:
		Techniques and Applications (DICTA)}, pages 1--9. IEEE, 2022.
	
	\bibitem{sanchez2023prospects}
	Thomas~W Sanchez, Hannah Shumway, Trey Gordner, and Theo Lim.
	\newblock The prospects of artificial intelligence in urban planning.
	\newblock {\em International Journal of Urban Sciences}, 27(2):179--194, 2023.
	
	\bibitem{ploennigs2023ai}
	Joern Ploennigs and Markus Berger.
	\newblock Ai art in architecture.
	\newblock {\em AI in Civil Engineering}, 2(1):8, 2023.
	
	\bibitem{chaillou2021ai}
	Stanislas Chaillou.
	\newblock Ai and architecture: An experimental perspective.
	\newblock In {\em The Routledge Companion to Artificial Intelligence in
		Architecture}, pages 420--441. Routledge, 2021.
	
	\bibitem{he2023revamping}
	Ziming He, Xiaomei Li, Ling Fan, and Harry~Jiannan Wang.
	\newblock Revamping interior design workflow through generative artificial
	intelligence.
	\newblock In {\em International Conference on Human-Computer Interaction},
	pages 607--613. Springer, 2023.
	
	\bibitem{hussein2023improving}
	GHADA~KHALED HUSSEIN et~al.
	\newblock Improving design efficiency using artificial intelligence: A study on
	the role of artificial intelligence in streamlining the interior design
	process.
	\newblock {\em International Design Journal}, 13(5):255--270, 2023.
	
	\bibitem{perez2017blurring}
	Rafael Iv{\'a}n~Pazos P{\'e}rez.
	\newblock {\em Blurring the boundaries between real and artificial in
		architecture and urban design through the use of artificial intelligence}.
	\newblock PhD thesis, Universidade da Coru{\~n}a, 2017.
	
	\bibitem{stigsenai}
	Mathias~Bank Stigsen, Alexandra Moisi, Shervin Rasoulzadeh, Kristina
	Schinegger, and Stefan Rutzinger.
	\newblock Ai diffusion as design vocabulary.
	
	\bibitem{kawar2022ddrm}
	Bahjat Kawar, Michael Elad, Stefano Ermon, and Jiaming Song.
	\newblock Denoising diffusion restoration models.
	\newblock {\em Advances in Neural Information Processing Systems},
	35:23593--23606, 2022.
	
	\bibitem{zhu2023denoising}
	Yuanzhi Zhu, Kai Zhang, Jingyun Liang, Jiezhang Cao, Bihan Wen, Radu Timofte,
	and Luc Van~Gool.
	\newblock Denoising diffusion models for plug-and-play image restoration.
	\newblock In {\em Proceedings of the IEEE/CVF Conference on Computer Vision and
		Pattern Recognition}, pages 1219--1229, 2023.
	
	\bibitem{saharia2022sr3}
	Chitwan Saharia, Jonathan Ho, William Chan, Tim Salimans, David~J Fleet, and
	Mohammad Norouzi.
	\newblock Image super-resolution via iterative refinement.
	\newblock {\em IEEE Transactions on Pattern Analysis and Machine Intelligence},
	45(4):4713--4726, 2022.
	
	\bibitem{mai2023ssif}
	Gengchen Mai, Ni~Lao, Weiwei Sun, Yuchi Ma, Jiaming Song, Chenlin Meng, Hongxu
	Ma, Jinmeng Rao, Ziyuan Li, and Stefano Ermon.
	\newblock Ssif: Learning continuous image representation for spatial-spectral
	super-resolution.
	\newblock {\em arXiv preprint arXiv:2310.00413}, 2023.
	
	\bibitem{kirillov2023sam}
	Alexander Kirillov, Eric Mintun, Nikhila Ravi, Hanzi Mao, Chloe Rolland, Laura
	Gustafson, Tete Xiao, Spencer Whitehead, Alexander~C Berg, Wan-Yen Lo, et~al.
	\newblock Segment anything.
	\newblock {\em arXiv preprint arXiv:2304.02643}, 2023.
	
	\bibitem{song2022objectstitch}
	Yizhi Song, Zhifei Zhang, Zhe Lin, Scott Cohen, Brian Price, Jianming Zhang,
	Soo~Ye Kim, and Daniel Aliaga.
	\newblock Objectstitch: Generative object compositing.
	\newblock {\em arXiv preprint arXiv:2212.00932}, 2022.
	
	\bibitem{singh2022paint2pix}
	Jaskirat Singh, Liang Zheng, Cameron Smith, and Jose Echevarria.
	\newblock Paint2pix: interactive painting based progressive image synthesis and
	editing.
	\newblock In {\em European Conference on Computer Vision}, pages 678--695.
	Springer, 2022.
	
	\bibitem{ho2022imagen}
	Jonathan Ho, William Chan, Chitwan Saharia, Jay Whang, Ruiqi Gao, Alexey
	Gritsenko, Diederik~P Kingma, Ben Poole, Mohammad Norouzi, David~J Fleet,
	et~al.
	\newblock Imagen video: High definition video generation with diffusion models.
	\newblock {\em arXiv preprint arXiv:2210.02303}, 2022.
	
	\bibitem{mustafa2023impact}
	Bahaa Mustafa.
	\newblock The impact of artificial intelligence on the graphic design industry.
	\newblock {\em resmilitaris}, 13(3):243--255, 2023.
	
	\bibitem{lu2023could}
	Yue Lu, Chao Guo, Yong Dou, Xingyuan Dai, and Fei-Yue Wang.
	\newblock Could chatgpt imagine: Content control for artistic painting
	generation via large language models.
	\newblock {\em Journal of Intelligent \& Robotic Systems}, 109(2):1--15, 2023.
	
	\bibitem{matthews2023destroy}
	Benjamin Matthews, Barrie Shannon, and Mark Roxburgh.
	\newblock Destroy all humans: The dematerialisation of the designer in an age
	of automation and its impact on graphic design—a literature review.
	\newblock {\em International Journal of Art \& Design Education},
	42(3):367--383, 2023.
	
	\bibitem{yuan2021font}
	Ye~Yuan, Wuyang Chen, Zhaowen Wang, Matthew Fisher, Zhifei Zhang, Zhangyang
	Wang, and Hailin Jin.
	\newblock Font completion and manipulation by cycling between multi-modality
	representations.
	\newblock {\em arXiv preprint arXiv:2108.12965}, 2021.
	
	\bibitem{yuan2020art}
	Ye~Yuan, Yasuaki Ito, and Koji Nakano.
	\newblock Art font image generation with conditional generative adversarial
	networks.
	\newblock In {\em 2020 Eighth International Symposium on Computing and
		Networking Workshops (CANDARW)}, pages 151--156. IEEE, 2020.
	
	\bibitem{zhang2023text2nerf}
	Jingbo Zhang, Xiaoyu Li, Ziyu Wan, Can Wang, and Jing Liao.
	\newblock Text2nerf: Text-driven 3d scene generation with neural radiance
	fields.
	\newblock {\em arXiv preprint arXiv:2305.11588}, 2023.
	
	\bibitem{yao2022learning}
	Chun-Han Yao, Jimei Yang, Duygu Ceylan, Yi~Zhou, Yang Zhou, and Ming-Hsuan
	Yang.
	\newblock Learning visibility for robust dense human body estimation.
	\newblock In {\em European Conference on Computer Vision}, pages 412--428.
	Springer, 2022.
	
	\bibitem{chen3using}
	Hung-Cheng Chen and Zhongwen Chen.
	\newblock Using chatgpt and midjourney to generate chinese landscape painting
	of tang poem ‘the difficult road to shu’.
	\newblock {\em International Journal of Social Sciences}, 3(2).
	
	\bibitem{romero2008art}
	Juan Romero, Juan~J Romero, and Penousal Machado.
	\newblock {\em The art of artificial evolution: A handbook on evolutionary art
		and music}.
	\newblock Springer Science \& Business Media, 2008.
	
	\bibitem{eiben2015introduction}
	Agoston~E Eiben and James~E Smith.
	\newblock {\em Introduction to evolutionary computing}.
	\newblock Springer, 2015.
	
	\bibitem{kim2020tivgan}
	Doyeon Kim, Donggyu Joo, and Junmo Kim.
	\newblock Tivgan: Text to image to video generation with step-by-step
	evolutionary generator.
	\newblock {\em IEEE Access}, 8:153113--153122, 2020.
	
	\bibitem{Tulyakov2018MoCoGAN}
	Sergey Tulyakov, Ming-Yu Liu, Xiaodong Yang, and Jan Kautz.
	\newblock {MoCoGAN}: Decomposing motion and content for video generation.
	\newblock In {\em IEEE Conference on Computer Vision and Pattern Recognition
		(CVPR)}, pages 1526--1535, 2018.
	
	\bibitem{weissenborn2019scaling}
	Dirk Weissenborn, Oscar T{\"a}ckstr{\"o}m, and Jakob Uszkoreit.
	\newblock Scaling autoregressive video models.
	\newblock {\em arXiv preprint arXiv:1906.02634}, 2019.
	
	\bibitem{le2021ccvs}
	Guillaume Le~Moing, Jean Ponce, and Cordelia Schmid.
	\newblock Ccvs: context-aware controllable video synthesis.
	\newblock {\em Advances in Neural Information Processing Systems},
	34:14042--14055, 2021.
	
	\bibitem{wu2022nuwa}
	Chenfei Wu, Jian Liang, Lei Ji, Fan Yang, Yuejian Fang, Daxin Jiang, and Nan
	Duan.
	\newblock N{\"u}wa: Visual synthesis pre-training for neural visual world
	creation.
	\newblock In {\em European conference on computer vision}, pages 720--736.
	Springer, 2022.
	
	\bibitem{ge2022long}
	Songwei Ge, Thomas Hayes, Harry Yang, Xi~Yin, Guan Pang, David Jacobs, Jia-Bin
	Huang, and Devi Parikh.
	\newblock Long video generation with time-agnostic vqgan and time-sensitive
	transformer.
	\newblock In {\em European Conference on Computer Vision}, pages 102--118.
	Springer, 2022.
	
	\bibitem{ho2022video}
	Jonathan Ho, Tim Salimans, Alexey Gritsenko, William Chan, Mohammad Norouzi,
	and David~J Fleet.
	\newblock Video diffusion models.
	\newblock {\em arXiv:2204.03458}, 2022.
	
	\bibitem{wu2023tune}
	Jay~Zhangjie Wu, Yixiao Ge, Xintao Wang, Stan~Weixian Lei, Yuchao Gu, Yufei
	Shi, Wynne Hsu, Ying Shan, Xiaohu Qie, and Mike~Zheng Shou.
	\newblock Tune-a-video: One-shot tuning of image diffusion models for
	text-to-video generation.
	\newblock In {\em Proceedings of the IEEE/CVF International Conference on
		Computer Vision}, pages 7623--7633, 2023.
	
	\bibitem{khachatryan2023text2video}
	Levon Khachatryan, Andranik Movsisyan, Vahram Tadevosyan, Roberto Henschel,
	Zhangyang Wang, Shant Navasardyan, and Humphrey Shi.
	\newblock Text2video-zero: Text-to-image diffusion models are zero-shot video
	generators.
	\newblock {\em arXiv preprint arXiv:2303.13439}, 2023.
	
	\bibitem{gpt4film}
	Chat gpt-4 in the film industry: Scriptwriting, editing, and more.
	\newblock
	\url{https://ts2.space/en/chat-gpt-4-in-the-film-industry-scriptwriting-editing-and-more/}.
	\newblock Accessed: 2023-09-25.
	
	\bibitem{huang2020movienet}
	Qingqiu Huang, Yu~Xiong, Anyi Rao, Jiaze Wang, and Dahua Lin.
	\newblock Movienet: A holistic dataset for movie understanding.
	\newblock In {\em The European Conference on Computer Vision (ECCV)}, 2020.
	
	\bibitem{Xiong_2019_ICCV}
	Yu~Xiong, Qingqiu Huang, Lingfeng Guo, Hang Zhou, Bolei Zhou, and Dahua Lin.
	\newblock A graph-based framework to bridge movies and synopses.
	\newblock In {\em The IEEE International Conference on Computer Vision (ICCV)},
	October 2019.
	
	\bibitem{GAIHolly}
	The impact of generative ai on hollywood and entertainment.
	\newblock
	\url{https://sloanreview.mit.edu/article/the-impact-of-generative-ai-on-hollywood-and-entertainment/}.
	\newblock Accessed: 2023-09-25.
	
	\bibitem{Huang_2018_CVPR}
	Qingqiu Huang, Yu~Xiong, and Dahua Lin.
	\newblock Unifying identification and context learning for person recognition.
	\newblock In {\em The IEEE Conference on Computer Vision and Pattern
		Recognition (CVPR)}, June 2018.
	
	\bibitem{huang2018person}
	Qingqiu Huang, Wentao Liu, and Dahua Lin.
	\newblock Person search in videos with one portrait through visual and temporal
	links.
	\newblock In {\em Proceedings of the European Conference on Computer Vision
		(ECCV)}, pages 425--441, 2018.
	
	\bibitem{xia2020online}
	Jiangyue Xia, Anyi Rao, Linning Xu, Qingqiu Huang, Jiangtao Wen, and Dahua Lin.
	\newblock Online multi-modal person search in videos.
	\newblock In {\em The European Conference on Computer Vision (ECCV)}, 2020.
	
	\bibitem{huang2018trailers}
	Qingqiu Huang, Yuanjun Xiong, Yu~Xiong, Yuqi Zhang, and Dahua Lin.
	\newblock From trailers to storylines: An efficient way to learn from movies.
	\newblock {\em arXiv preprint arXiv:1806.05341}, 2018.
	
	\bibitem{rao2020unified}
	Anyi Rao, Jiaze Wang, Linning Xu, Xuekun Jiang, Qingqiu Huang, Bolei Zhou, and
	Dahua Lin.
	\newblock A unified framework for shot type classification based on subject
	centric lens.
	\newblock In {\em The European Conference on Computer Vision (ECCV)}, 2020.
	
	\bibitem{rao2020local}
	Anyi Rao, Linning Xu, Yu~Xiong, Guodong Xu, Qingqiu Huang, Bolei Zhou, and
	Dahua Lin.
	\newblock A local-to-global approach to multi-modal movie scene segmentation.
	\newblock In {\em Proceedings of the IEEE/CVF Conference on Computer Vision and
		Pattern Recognition}, pages 10146--10155, 2020.
	
	\bibitem{aionmedia}
	The effects of artificial intelligence on media and communication.
	\newblock
	\url{https://www.linkedin.com/pulse/effects-artificial-intelligence-media-communication-mayowa-lateef/}.
	\newblock Accessed: 2023-09-25.
	
	\bibitem{ailocalnews}
	Ai and the future of local tv news.
	\newblock Accessed: 2023-09-25.
	
	\bibitem{zhang2023video}
	Hang Zhang, Xin Li, and Lidong Bing.
	\newblock Video-llama: An instruction-tuned audio-visual language model for
	video understanding.
	\newblock {\em arXiv preprint arXiv:2306.02858}, 2023.
	
	\bibitem{li2017video}
	Bochen Li, Karthik Dinesh, Gaurav Sharma, and Zhiyao Duan.
	\newblock Video-based vibrato detection and analysis for polyphonic string
	music.
	\newblock In {\em ISMIR}, pages 123--130, 2017.
	
	\bibitem{li2018skeleton}
	Bochen Li, Akira Maezawa, and Zhiyao Duan.
	\newblock Skeleton plays piano: Online generation of pianist body movements
	from midi performance.
	\newblock In {\em ISMIR}, pages 218--224, 2018.
	
	\bibitem{li2017see}
	Bochen Li, Karthik Dinesh, Zhiyao Duan, and Gaurav Sharma.
	\newblock See and listen: Score-informed association of sound tracks to players
	in chamber music performance videos.
	\newblock In {\em 2017 IEEE International Conference on Acoustics, Speech and
		Signal Processing (ICASSP)}, pages 2906--2910. IEEE, 2017.
	
	\bibitem{li2019online}
	Bochen Li, Karthik Dinesh, Chenliang Xu, Gaurav Sharma, and Zhiyan Duan.
	\newblock Online audio-visual source association for chamber music
	performances.
	\newblock {\em Transactions of the International Society for Music Information
		Retrieval}, 2(1), 2019.
	
	\bibitem{heydari2023singnet}
	Mojtaba Heydari, Ju-Chiang Wang, and Zhiyao Duan.
	\newblock Singnet: a real-time singing voice beat and downbeat tracking system.
	\newblock In {\em ICASSP 2023-2023 IEEE International Conference on Acoustics,
		Speech and Signal Processing (ICASSP)}, pages 1--5. IEEE, 2023.
	
	\bibitem{sargent2014segmentation}
	Gabriel Sargent, Pierre Hanna, and Henri Nicolas.
	\newblock Segmentation of music video streams in music pieces through
	audio-visual analysis.
	\newblock In {\em 2014 IEEE International Conference on Acoustics, Speech and
		Signal Processing (ICASSP)}, pages 724--728. IEEE, 2014.
	
	\bibitem{killian2001effect}
	Janice~N Killian.
	\newblock The effect of audio, visual and audio-visual performance on
	perception of musical content.
	\newblock {\em Bulletin of the Council for Research in Music Education}, pages
	77--87, 2001.
	
	\bibitem{li2018creating}
	Bochen Li, Xinzhao Liu, Karthik Dinesh, Zhiyao Duan, and Gaurav Sharma.
	\newblock Creating a multitrack classical music performance dataset for
	multimodal music analysis: Challenges, insights, and applications.
	\newblock {\em IEEE Transactions on Multimedia}, 21(2):522--535, 2018.
	
	\bibitem{thompson2008audio}
	William~Forde Thompson, Frank~A Russo, and Lena Quinto.
	\newblock Audio-visual integration of emotional cues in song.
	\newblock {\em Cognition and Emotion}, 22(8):1457--1470, 2008.
	
	\bibitem{duan2019performances}
	Zhiyao Duan, Slim Essid, Cynthia C.~S. Liem, Gaël Richard, and Gaurav Sharma.
	\newblock Audio-visual analysis of music performances.
	\newblock {\em IEEE Signal Processing Magazine}, 36(1):63--73, 2019.
	
	\bibitem{bellaiche2023humans}
	Lucas Bellaiche, Rohin Shahi, Martin~Harry Turpin, Anya Ragnhildstveit, Shawn
	Sprockett, Nathaniel Barr, Alexander Christensen, and Paul Seli.
	\newblock Humans versus ai: whether and why we prefer human-created compared to
	ai-created artwork.
	\newblock {\em Cognitive Research: Principles and Implications}, 8(1):1--22,
	2023.
	
	\bibitem{ji2023survey}
	Ziwei Ji, Nayeon Lee, Rita Frieske, Tiezheng Yu, Dan Su, Yan Xu, Etsuko Ishii,
	Ye~Jin Bang, Andrea Madotto, and Pascale Fung.
	\newblock Survey of hallucination in natural language generation.
	\newblock {\em ACM Computing Surveys}, 55(12):1--38, 2023.
	
	\bibitem{borji2023categorical}
	Ali Borji.
	\newblock A categorical archive of chatgpt failures.
	\newblock {\em arXiv preprint arXiv:2302.03494}, 2023.
	
	\bibitem{handsfigers}
	Why does ai art screw up hands and fingers?
	\newblock
	\url{https://www.britannica.com/topic/Why-does-AI-art-screw-up-hands-and-fingers-2230501}.
	\newblock Accessed: 2023-09-25.
	
	\bibitem{lin2004rouge}
	Chin-Yew Lin.
	\newblock Rouge: A package for automatic evaluation of summaries.
	\newblock In {\em Text summarization branches out}, pages 74--81, 2004.
	
	\bibitem{papineni2002bleu}
	Kishore Papineni, Salim Roukos, Todd Ward, and Wei-Jing Zhu.
	\newblock Bleu: a method for automatic evaluation of machine translation.
	\newblock In {\em Proceedings of the 40th annual meeting of the Association for
		Computational Linguistics}, pages 311--318, 2002.
	
	\bibitem{goodrich2019assessing}
	Ben Goodrich, Vinay Rao, Peter~J Liu, and Mohammad Saleh.
	\newblock Assessing the factual accuracy of generated text.
	\newblock In {\em proceedings of the 25th ACM SIGKDD international conference
		on knowledge discovery \& data mining}, pages 166--175, 2019.
	
	\bibitem{evans2021truthful}
	Owain Evans, Owen Cotton-Barratt, Lukas Finnveden, Adam Bales, Avital Balwit,
	Peter Wills, Luca Righetti, and William Saunders.
	\newblock Truthful ai: Developing and governing ai that does not lie.
	\newblock {\em arXiv preprint arXiv:2110.06674}, 2021.
	
	\bibitem{guo2023aigc}
	Danhuai Guo, Huixuan Chen, Ruoling Wu, and Yangang Wang.
	\newblock Aigc challenges and opportunities related to public safety: A case
	study of chatgpt.
	\newblock {\em Journal of Safety Science and Resilience}, 2023.
	
	\bibitem{rao2023building}
	Jinmeng Rao, Song Gao, Gengchen Mai, and Krzysztof Janowicz.
	\newblock Building privacy-preserving and secure geospatial artificial
	intelligence foundation models.
	\newblock {\em arXiv preprint arXiv:2309.17319}, 2023.
	
	\bibitem{zhou2023synthetic}
	Jiawei Zhou, Yixuan Zhang, Qianni Luo, Andrea~G Parker, and Munmun
	De~Choudhury.
	\newblock Synthetic lies: Understanding ai-generated misinformation and
	evaluating algorithmic and human solutions.
	\newblock In {\em Proceedings of the 2023 CHI Conference on Human Factors in
		Computing Systems}, pages 1--20, 2023.
	
	\bibitem{agrawal2021survey}
	Ronak Agrawal and Dilip~Kumar Sharma.
	\newblock A survey on video-based fake news detection techniques.
	\newblock In {\em 2021 8th International Conference on Computing for
		Sustainable Global Development (INDIACom)}, pages 663--669. IEEE, 2021.
	
	\bibitem{fakeNewsImage}
	Popular image generators accept 85% of fake news prompts.
	\newblock
	\url{https://aibusiness.com/responsible-ai/popular-image-generators-accept-85-of-fake-news-prompts}.
	\newblock Accessed: 2023-09-25.
	
	\bibitem{trump}
	Fake trump arrest photos: How to spot an ai-generated image.
	\newblock \url{https://www.bbc.com/news/world-us-canada-65069316}.
	\newblock Accessed: 2023-09-25.
	
	\bibitem{biden}
	Pictures of joe biden and vp celebrating trump indictment are fake.
	\newblock \url{https://factcheck.afp.com/doc.afp.com.33CK93W}.
	\newblock Accessed: 2023-09-25.
	
	\bibitem{deepfakeVideo}
	Deepfake scams have arrived: Fake videos spread on facebook, tiktok and
	youtube.
	\newblock
	\url{https://www.nbcnews.com/tech/tech-news/deepfake-scams-arrived-fake-videos-spread-facebook-tiktok-youtube-rcna101415}.
	\newblock Accessed: 2023-09-25.
	
	\bibitem{tucker2019newest}
	Patrick Tucker.
	\newblock The newest ai-enabled weapon: Deep-faking photos of the earth.
	\newblock {\em Defense One}, (13):03, 2019.
	
	\bibitem{zhao2021deepfakegeography}
	Bo~Zhao, Shaozeng Zhang, Chunxue Xu, Yifan Sun, and Chengbin Deng.
	\newblock Deep fake geography? when geospatial data encounter artificial
	intelligence.
	\newblock {\em Cartography and Geographic Information Science}, 48(4):338--352,
	2021.
	
	\bibitem{caputo2013going}
	Deanna~D Caputo, Shari~Lawrence Pfleeger, Jesse~D Freeman, and M~Eric Johnson.
	\newblock Going spear phishing: Exploring embedded training and awareness.
	\newblock {\em IEEE security \& privacy}, 12(1):28--38, 2013.
	
	\bibitem{hazell2023large}
	Julian Hazell.
	\newblock Large language models can be used to effectively scale spear phishing
	campaigns.
	\newblock {\em arXiv preprint arXiv:2305.06972}, 2023.
	
	\bibitem{manyam2022artificial}
	Sowjanya Manyam.
	\newblock Artificial intelligence's impact on social engineering attacks.
	\newblock 2022.
	
	\bibitem{roy2023generating}
	Sayak~Saha Roy, Krishna~Vamsi Naragam, and Shirin Nilizadeh.
	\newblock Generating phishing attacks using chatgpt.
	\newblock {\em arXiv preprint arXiv:2305.05133}, 2023.
	
	\bibitem{fang2023bias}
	Xiao Fang, Shangkun Che, Minjia Mao, Hongzhe Zhang, Ming Zhao, and Xiaohang
	Zhao.
	\newblock Bias of ai-generated content: An examination of news produced by
	large language models.
	\newblock {\em arXiv preprint arXiv:2309.09825}, 2023.
	
	\bibitem{biasMarketingContent}
	Overcoming algorithmic gender bias in ai-generated marketing content.
	\newblock
	\url{https://www.forbes.com/sites/forbescommunicationscouncil/2023/07/25/overcoming-algorithmic-gender-bias-in-ai-generated-marketing-content/?sh=518d4fa11639}.
	\newblock Accessed: 2023-09-25.
	
	\bibitem{manvi2023geollm}
	Rohin Manvi, Samar Khanna, Gengchen Mai, Marshall Burke, David Lobell, and
	Stefano Ermon.
	\newblock Geollm: Extracting geospatial knowledge from large language models.
	\newblock {\em arXiv preprint arXiv:2310.06213}, 2023.
	
	\bibitem{faisal2022geographicbias}
	Fahim Faisal and Antonios Anastasopoulos.
	\newblock Geographic and geopolitical biases of language models.
	\newblock {\em arXiv preprint arXiv:2212.10408}, 2022.
	
	\bibitem{tang2023science}
	Ruixiang Tang, Yu-Neng Chuang, and Xia Hu.
	\newblock The science of detecting llm-generated texts.
	\newblock {\em arXiv preprint arXiv:2303.07205}, 2023.
	
	\bibitem{ippolito2019automatic}
	Daphne Ippolito, Daniel Duckworth, Chris Callison-Burch, and Douglas Eck.
	\newblock Automatic detection of generated text is easiest when humans are
	fooled.
	\newblock {\em arXiv preprint arXiv:1911.00650}, 2019.
	
	\bibitem{davidson2017automated}
	Thomas Davidson, Dana Warmsley, Michael Macy, and Ingmar Weber.
	\newblock Automated hate speech detection and the problem of offensive
	language.
	\newblock In {\em Proceedings of the international AAAI conference on web and
		social media}, volume~11, pages 512--515, 2017.
	
	\bibitem{zampieri2019semeval}
	Marcos Zampieri, Shervin Malmasi, Preslav Nakov, Sara Rosenthal, Noura Farra,
	and Ritesh Kumar.
	\newblock Semeval-2019 task 6: Identifying and categorizing offensive language
	in social media (offenseval).
	\newblock In {\em Proceedings of the 13th International Workshop on Semantic
		Evaluation}, pages 75--86, 2019.
	
	\bibitem{deshpande2023toxicity}
	Ameet Deshpande, Vishvak Murahari, Tanmay Rajpurohit, Ashwin Kalyan, and
	Karthik Narasimhan.
	\newblock Toxicity in chatgpt: Analyzing persona-assigned language models.
	\newblock {\em arXiv preprint arXiv:2304.05335}, 2023.
	
	\bibitem{gehman2020realtoxicityprompts}
	Samuel Gehman, Suchin Gururangan, Maarten Sap, Yejin Choi, and Noah~A Smith.
	\newblock Realtoxicityprompts: Evaluating neural toxic degeneration in language
	models.
	\newblock {\em arXiv e-prints}, pages arXiv--2009, 2020.
	
	\bibitem{ngo2021mitigating}
	Helen Ngo, Cooper Raterink, Jo{\~a}o~GM Ara{\'u}jo, Ivan Zhang, Carol Chen,
	Adrien Morisot, and Nicholas Frosst.
	\newblock Mitigating harm in language models with conditional-likelihood
	filtration.
	\newblock {\em arXiv preprint arXiv:2108.07790}, 2021.
	
	\bibitem{prabhumoye2023adding}
	Shrimai Prabhumoye, Mostofa Patwary, Mohammad Shoeybi, and Bryan Catanzaro.
	\newblock Adding instructions during pretraining: Effective way of controlling
	toxicity in language models.
	\newblock In {\em Proceedings of the 17th Conference of the European Chapter of
		the Association for Computational Linguistics}, pages 2628--2643, 2023.
	
	\bibitem{wang2022exploring}
	Boxin Wang, Wei Ping, Chaowei Xiao, Peng Xu, Mostofa Patwary, Mohammad Shoeybi,
	Bo~Li, Anima Anandkumar, and Bryan Catanzaro.
	\newblock Exploring the limits of domain-adaptive training for detoxifying
	large-scale language models.
	\newblock {\em Advances in Neural Information Processing Systems},
	35:35811--35824, 2022.
	
	\bibitem{ouyang2022training}
	Long Ouyang, Jeffrey Wu, Xu~Jiang, Diogo Almeida, Carroll Wainwright, Pamela
	Mishkin, Chong Zhang, Sandhini Agarwal, Katarina Slama, Alex Ray, et~al.
	\newblock Training language models to follow instructions with human feedback.
	\newblock {\em Advances in Neural Information Processing Systems},
	35:27730--27744, 2022.
	
	\bibitem{gururangan2020don}
	Suchin Gururangan, Ana Marasovi{\'c}, Swabha Swayamdipta, Kyle Lo, Iz~Beltagy,
	Doug Downey, and Noah~A Smith.
	\newblock Don’t stop pretraining: Adapt language models to domains and tasks.
	\newblock In {\em Proceedings of the 58th Annual Meeting of the Association for
		Computational Linguistics}, pages 8342--8360, 2020.
	
	\bibitem{he2023you}
	Xinlei He, Savvas Zannettou, Yun Shen, and Yang Zhang.
	\newblock You only prompt once: On the capabilities of prompt learning on large
	language models to tackle toxic content.
	\newblock {\em arXiv e-prints}, pages arXiv--2308, 2023.
	
	\bibitem{dathathri2019plug}
	Sumanth Dathathri, Andrea Madotto, Janice Lan, Jane Hung, Eric Frank, Piero
	Molino, Jason Yosinski, and Rosanne Liu.
	\newblock Plug and play language models: A simple approach to controlled text
	generation.
	\newblock In {\em International Conference on Learning Representations}, 2019.
	
	\bibitem{liu2021dexperts}
	Alisa Liu, Maarten Sap, Ximing Lu, Swabha Swayamdipta, Chandra Bhagavatula,
	Noah~A Smith, and Yejin Choi.
	\newblock Dexperts: Decoding-time controlled text generation with experts and
	anti-experts.
	\newblock In {\em Proceedings of the 59th Annual Meeting of the Association for
		Computational Linguistics and the 11th International Joint Conference on
		Natural Language Processing (Volume 1: Long Papers)}, pages 6691--6706, 2021.
	
\end{thebibliography}

\end{document}